\documentclass[letterpaper, 10 pt, journal, twoside]{ieeetran}

\IEEEoverridecommandlockouts

\usepackage{cite}
\usepackage{amsmath,amssymb,amsfonts}
\usepackage{graphicx}
\usepackage{textcomp}
\usepackage{xcolor}
\usepackage{booktabs}
\usepackage{balance}

\usepackage{siunitx}
\usepackage[super]{nth}
\usepackage[export]{adjustbox}

\usepackage{caption}
\usepackage{subcaption}

\usepackage{enumitem}
\setlist{leftmargin=15pt}

\usepackage{titlesec}
\titlespacing*{\paragraph}
{0pt}{3.25ex plus 1ex minus .2ex}{1.5ex plus .2ex}
\titlespacing*{\section}
{0em}{0.5em}{0.3em}
\titlespacing*{\subsection}
{0em}{0.5em}{0.3em}
\titlespacing*{\subsubsection}
{0em}{0.5em}{0.3em}

\usepackage{algorithm}
\usepackage[noend]{algpseudocode}
\makeatletter
\def\BState{\State\hskip-\ALG@thistlm}
\makeatother
\newcommand{\function}[1]{\textbf{#1}}

\newcommand{\vk}[1]{#1}
\newcommand{\vkcaption}[1]{#1}

\def\BibTeX{{\rm B\kern-.05em{\sc i\kern-.025em b}\kern-.08em
    T\kern-.1667em\lower.7ex\hbox{E}\kern-.125emX}}

\usepackage{hyperref} 
\hypersetup{
  colorlinks,
  citecolor=black,
  filecolor=black,
  linkcolor=black,
  urlcolor=black,
  pdfauthor={},
  pdfsubject={},
  pdftitle={}
}

\usepackage{scalerel}
\usepackage{tikz}
\usetikzlibrary{svg.path}
\definecolor{orcidlogocol}{HTML}{A6CE39}
\tikzset{
  orcidlogo/.pic={
    \fill[orcidlogocol] svg{M256,128c0,70.7-57.3,128-128,128C57.3,256,0,198.7,0,128C0,57.3,57.3,0,128,0C198.7,0,256,57.3,256,128z};
    \fill[white] svg{M86.3,186.2H70.9V79.1h15.4v48.4V186.2z}
    svg{M108.9,79.1h41.6c39.6,0,57,28.3,57,53.6c0,27.5-21.5,53.6-56.8,53.6h-41.8V79.1z M124.3,172.4h24.5c34.9,0,42.9-26.5,42.9-39.7c0-21.5-13.7-39.7-43.7-39.7h-23.7V172.4z}
    svg{M88.7,56.8c0,5.5-4.5,10.1-10.1,10.1c-5.6,0-10.1-4.6-10.1-10.1c0-5.6,4.5-10.1,10.1-10.1C84.2,46.7,88.7,51.3,88.7,56.8z};
  }
}

\newcommand\orcidicon[1]{\href{https://orcid.org/#1}{\mbox{\scalerel*{
        \begin{tikzpicture}[yscale=-1,transform shape]
          \pic{orcidlogo};
        \end{tikzpicture}
}{|}}}}

\definecolor{light_blue}{rgb}{0.1, 0.1, 0.8}

\begin{document}

\newcommand{\PREPRINTYEAR}{2022}
\newcommand{\PREPRINTPUBLISHER}{IEEE}

\onecolumn
\pagenumbering{gobble}
{
  \topskip0pt
  \vspace*{\fill}
  \centering
  \LARGE{%
    \copyright{} \PREPRINTYEAR~\PREPRINTPUBLISHER\\\vspace{1cm}
	Personal use of this material is permitted.
	Permission from \PREPRINTPUBLISHER~must be obtained for all other uses, in any current or future media, including reprinting or republishing this material for advertising or promotional purposes, creating new collective works, for resale or redistribution to servers or lists, or reuse of any copyrighted component of this work in other works.}
	\vspace*{\fill}
}

\twocolumn 
\pagenumbering{arabic}

\markboth{\copyright{} \PREPRINTPUBLISHER, \PREPRINTYEAR. Accepted to IEEE Robotics and Automation Letters (RA-L). DOI: \href{https://doi.org/10.1109/LRA.2022.3195194}{10.1109/LRA.2022.3195194}}{\copyright{} \PREPRINTPUBLISHER, \PREPRINTYEAR. Accepted to IEEE Robotics and Automation Letters (RA-L). DOI: \href{https://doi.org/10.1109/LRA.2022.3195194}{10.1109/LRA.2022.3195194}}

\def\map{SphereMap}
\def\segmap{LTVMap}
\def\paper{letter}
\def\mytitle{\map{}: Dynamic Multi-Layer Graph Structure for Rapid Safety-Aware UAV Planning}

\title{\mytitle{}}

\author{
  Tom\'{a}\v{s} Musil$^{\orcidicon{0000-0002-9421-6544}}$, 
  Mat\v{e}j Petrl\'{i}k$^{\orcidicon{0000-0002-5337-9558}}$,
  Martin Saska$^{\orcidicon{0000-0001-7106-3816}}$%
\thanks{%
  Manuscript received February 24, 2022; Revised June 20, 2022; Accepted July 18, 2022.
  This paper was recommended for publication by Editor Tamim Asfour upon evaluation of the Associate Editor and Reviewers’ comments.}
  \thanks{%
    \vkcaption{
    This work was supported
    by the Defense Advanced Research Projects Agency (DARPA),
    by the Czech Science Foundation (GA\v{C}R) under research project no. 20-29531S,
    by TA\v{C}R, project no. FW03010020,
    by project no. DG18P02OVV069 in program NAKI II,
    by CTU grant no. SGS20/174/OHK3/3T/13, and
    by OP VVV funded project CZ.02.1.01/0.0/0.0/16 019/0000765 "Research Center for Informatics".
  }
    }
  \thanks{%
    Authors are with the Department of Cybernetics, Faculty of Electrical Engineering, Czech Technical University in Prague, 166 36 Prague 6, {\tt\footnotesize\{\href{mailto:tomas.musil@fel.cvut.cz}{tomas.musil}|\href{mailto:matej.petrlik@fel.cvut.cz}{matej.petrlik}|\href{mailto:martin.saska@fel.cvut.cz}{martin.saska}\}@fel.cvut.cz}
}
  \thanks{Digital Object Identifier (DOI): see top of this page.}
}

\maketitle

\begin{abstract}
  A flexible topological representation consisting of a two-layer graph structure built on-board an Unmanned Aerial Vehicle (UAV) by continuously filling the free space of an occupancy map with intersecting spheres is proposed in this \paper{}.
  Most state-of-the-art planning methods find the shortest paths while keeping the UAV at a pre-defined distance from obstacles.
  Planning over the proposed structure reaches this pre-defined distance only when necessary, maintaining a safer distance otherwise, while also being orders of magnitude faster than other state-of-the-art methods.
  Furthermore, we demonstrate how this graph representation can be converted into a lightweight shareable topological-volumetric map of the environment, which enables decentralized multi-robot cooperation.
  The proposed approach was successfully validated in several kilometers of real subterranean environments, such as caves, devastated industrial buildings, and in the harsh and complex setting of the final event of the DARPA SubT Challenge, which aims to mimic the conditions of real search and rescue missions as closely as possible, and where our approach achieved the \nth{2} place in the virtual track.
\end{abstract}


\section*{Multimedia Materials}
\label{sec:multimedia_materials}
The paper is supported by the multimedia materials available at \href{http://mrs.felk.cvut.cz/papers/ral2022spheremap}{mrs.felk.cvut.cz/papers/ral2022spheremap}.

\section{Introduction}

\begin{figure}[t]
  \centering
  \includegraphics[width=1.0\linewidth]{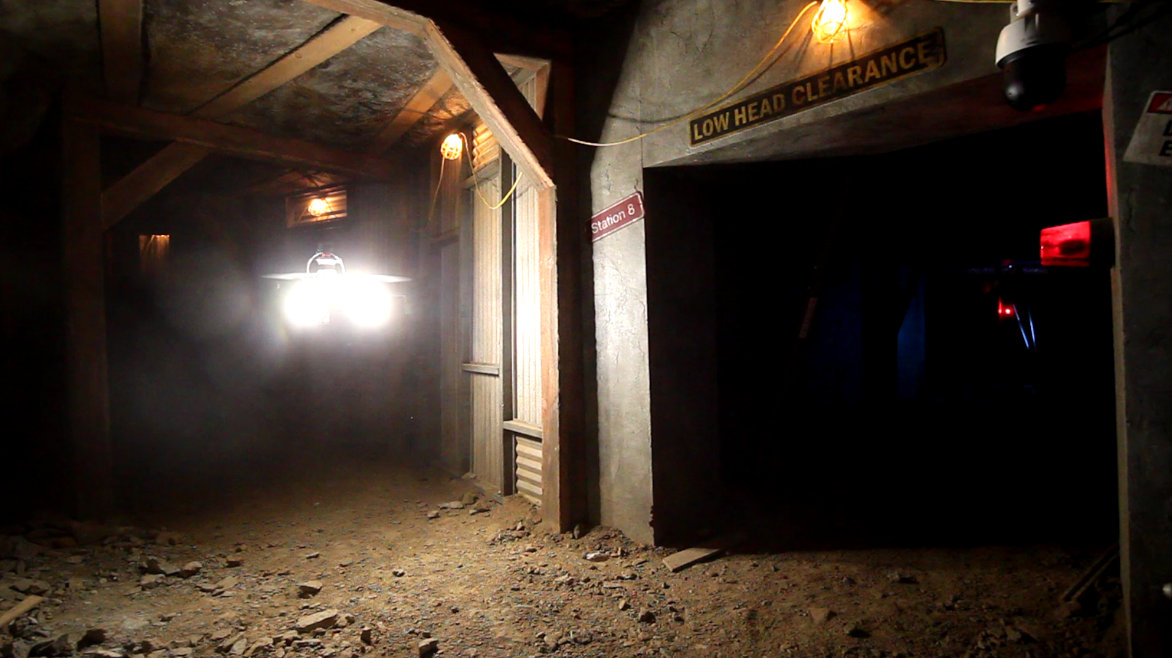}
  \adjincludegraphics[width=0.32\linewidth, trim={{0.27\width} {0.35\height} {0.20\width} {0.10\height}}, clip]{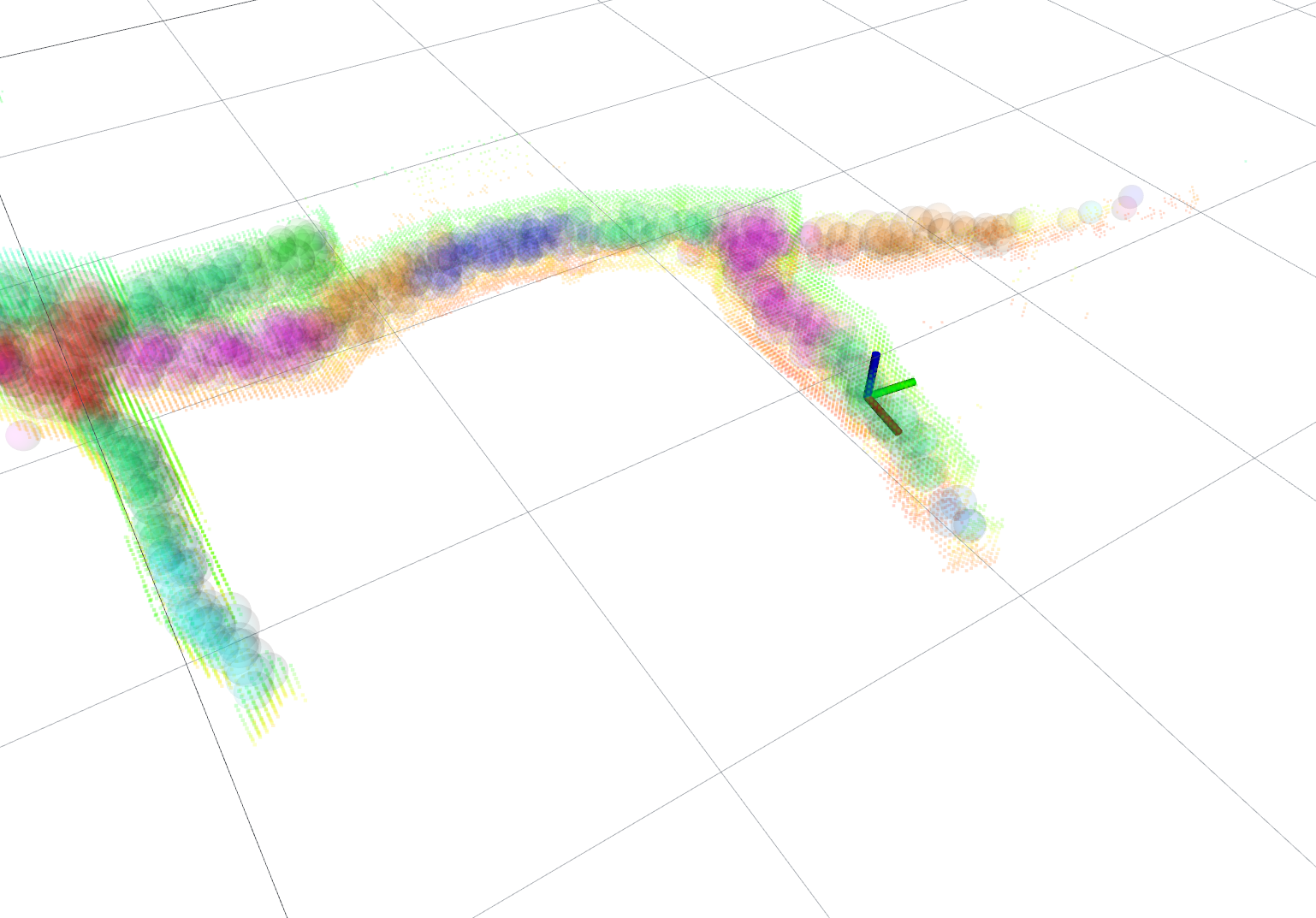}
  \adjincludegraphics[width=0.32\linewidth, trim={{0.27\width} {0.35\height} {0.20\width} {0.10\height}}, clip]{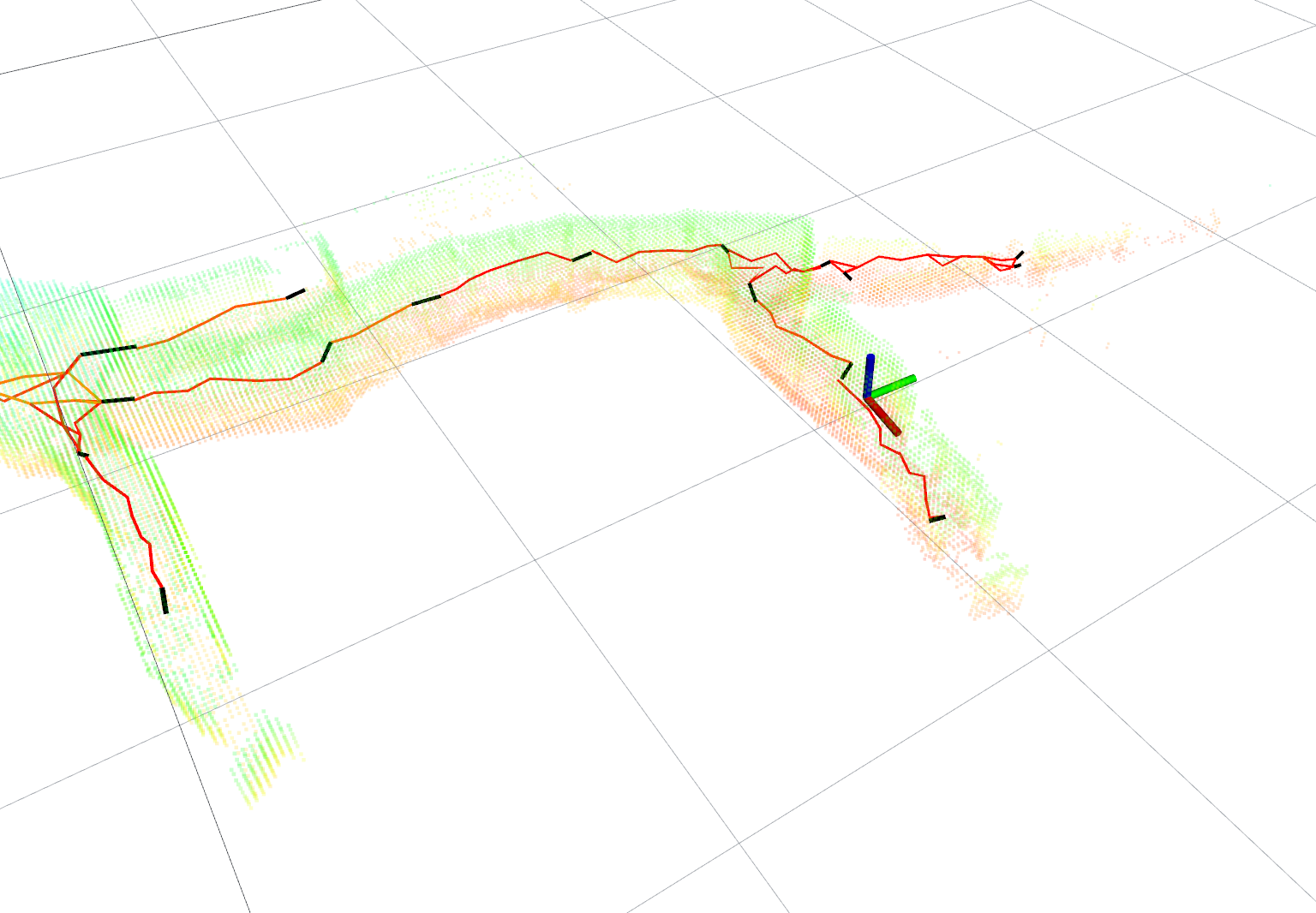}
  \adjincludegraphics[width=0.32\linewidth, trim={{0.27\width} {0.35\height} {0.20\width} {0.10\height}}, clip]{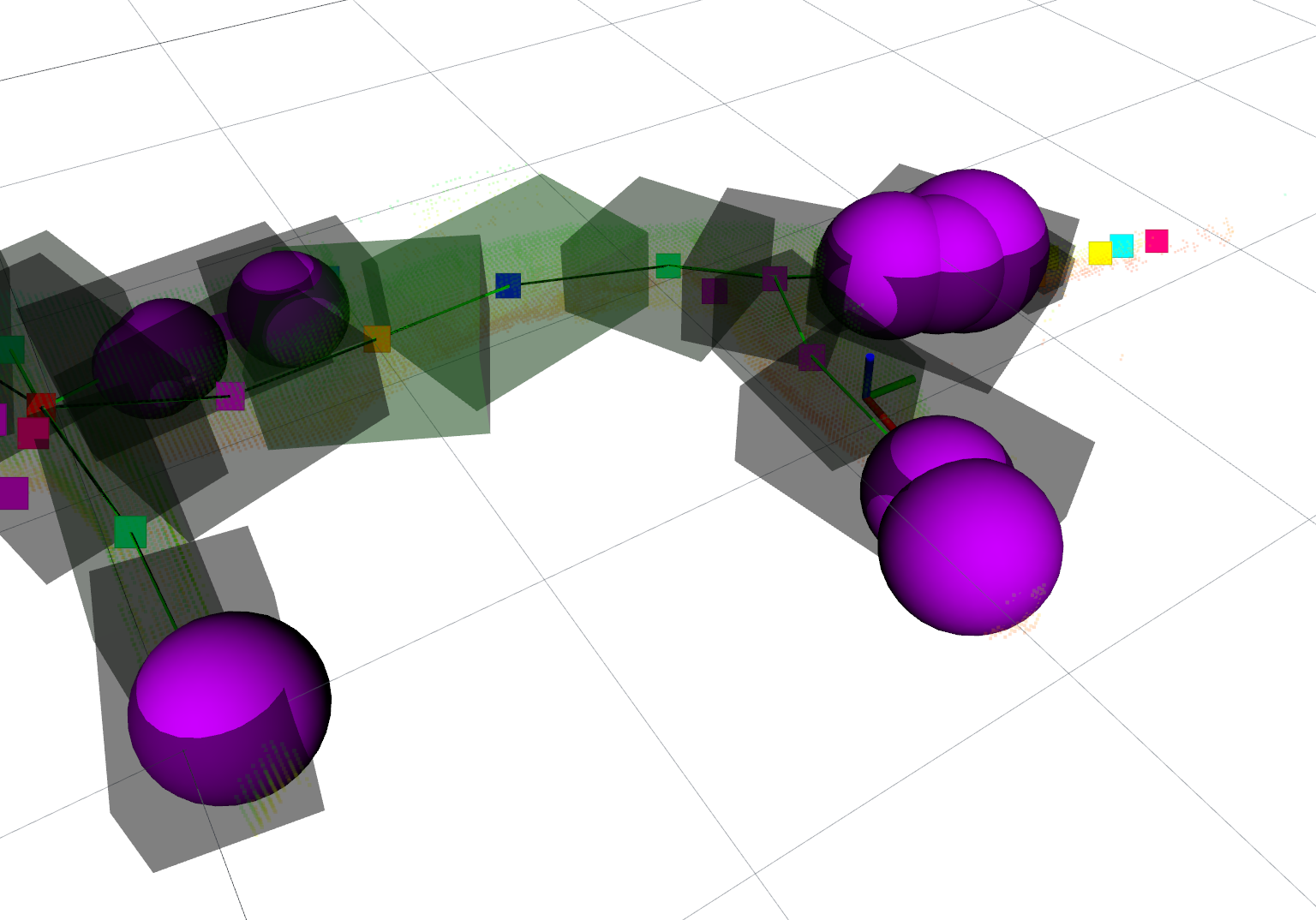}
  \caption{The \map{} created during the final event of DARPA SubT Challenge in the tunnel section of the course.
  The free space is filled with spheres (left), forming an undirected graph with spherical centers as vertices.
  Middle image shows portals (black) between pairs of roughly convex regions with cached intra-region paths (red).
  The bounding boxes of the \segmap{} is shown on the right with darker colors representing less explored/inspected regions, and purple spheres marking exploration goals.
  }
  \label{fig:intro_img}
\end{figure}

\IEEEPARstart{T}{hanks} to the rising affordability and rapid development of onboard software, autonomous UAVs are increasingly being deployed in Search and Rescue (SAR) missions \cite{delmerico2019current} and inspection of industrial \cite{carvalho2020uav} or historical \cite{petracek2020ral} structures.
The high mobility of UAVs makes them an ideal tool for first-responders and rescue teams.
A quick flight through an unknown and potentially dangerous environment provides the operator with quick situational awareness that allows better decision making and avoids risking human lives when exploring uncharted environment with unknown safety conditions.

Systems for accomplishing autonomous tasks by UAVs are available in literature \cite{baca2021mrs, sanchez2016aerostack} and can be deployed in various scenarios, such as inspection \cite{kratky2021aerialfilming}, firefighting \cite{spurny2021autonomous, quenzel2021autonomous} and radiation localization \cite{baca2021icuas}.
These applications often rely on GNSS for UAV localization and additional sensors for mostly simple reactive behavior.
On the other hand, the navigation of UAVs in large unstructured environments without reliable GNSS data is still an open topic, one in which advancements have been recently accelerated thanks to the \vk{DARPA SubT} tunnel circuit \cite{petrlik2020robust}, urban circuit \cite{kratky2021exploration}, and cave circuit \cite{petracek2021caves}.

Finding a collision-free path through any environment requires a representation of the free space and obstacles.
The commonly employed volumetric occupancy grids based on octree data structures (e.g. the OctoMap \cite{octomap} or UFOMap \cite{duberg2020ufomap} implementations) encode this information as the states of cells.
Even though octree-based occupancy maps are memory efficient and allow fast access times, naive planning on an occupancy grid may be unfeasible in real time for long distance planning (global navigation), especially with complex topology typical for subterranean environments.
Timeout-based workarounds exist \cite{kratky2021exploration}, which output the path with an endpoint of the lowest Euclidean distance from the initial goal. 
This approach avoids long idle times while waiting for the plan, but might prevent reaching the goal altogether in cases of non-trivial topology.
Many approaches to the global navigation problem utilize building a map exploiting the topology of the environment for rapid path planning, such as \cite{skeletons, topomap, convex_exploration}.
However, most of these methods build the topological map after an initial run-through of the environment and not incrementally throughout the mission. 
Thus they cannot be deployed on a first responder UAV into a previously unmapped environment, which is a typical requirement for SAR scenarios. 
Those that are built incrementally often do not incorporate obstacle clearance information into the map for safety-aware planning.
As such, even those methods cannot be used when the UAV needs to quickly decide whether to explore a nearby narrow passage with a high risk of collision, or to explore a farther goal for which a safer path exists.

This \paper{} presents an incrementally built graph structure (called \map{}),  consisting of a sparse topological graph of roughly convex regions of free space, and an underlying dense graph of intersecting spheres.
This structure can be built continually on-board a UAV during the mission, and allow it to make decisions about mission goals, weighing the safety of paths, their length, and remaining flight time in real time.
Furthermore, the structure can generate a lightweight representation (called \segmap{}) of the environment consisting of a graph of convex shapes representing the environment's volume to share with other robots or human operators at any point.  

\subsection{Related works}
A lightweight topological map representation designed for quick path planning was presented in \cite{topomap}.
In this work, the authors grow convex clusters of free space from a Euclidean Signed Distance Field (ESDF) created offline after one run of the robot through the environment.
The obtained maps are then used for rapid path planning in the environment.
Since the clusters are convex, the path is planned on a navigation graph where the vertices are points (called portals) in which the clusters coincide.
Edges are straight-line paths between two portals through a cluster.
The approach requires a fixed voxel size for the clusters, which can pose problems when the environment has both very narrow, and large open areas.
Moreover, the offline creation of an ESDF prevents exploratory deployment in an unknown environment and cannot handle dynamically changing environments.

Another approach of creating a sparse graph representation of the environment is found in \cite{skeletons}.
A 3D Voronoi diagram is created from an ESDF map and converted to a sparse graph, which enables planning that is orders of magnitude faster than A* or RRT* \cite{karaman2011sampling} on the ESDF.
The sparse graph is, however, not built dynamically, but only after an initial exploration mission. Clearance information is also not incorporated into the graph, though the authors state this is to be done in their future work.

A popular method of constructing a topological graph for the purpose of autonomous exploration, called GBPlanner, was presented in \cite{gbplanner}.
In this approach, a global graph is incrementally built during the mission and is used for homing and repositioning the UAV to new frontiers when a dead end is reached.
The downside of this approach is that it does not enable easy querying of a path between any two points in the occupancy map, and also that the graph is built to provide the shortest path home. This does not allow for weighing path length \vk{vs.} path safety when planning towards new frontiers.

A similar approach was presented \vk{in~\cite{convex_exploration}}, where the authors incrementally build a sparse topological graph of coverage points from which they cast rays through an occupancy map and create a convex polyhedron at each coverage point.
These coverage points are also merged if a centroid of a polyhedron is contained inside another polyhedron.
This is an incremental and very lightweight topological representation, but it does not contain implicit clearance information, and the authors present path planning based only on distance.

A technique for safety-aware path planning was presented in \cite{spartan}, where the authors construct the path optimization criterion to contain information about the distance from obstacles up to a maximum distance $d_{max}$ at a given point on the path.
In this approach, the authors solve the optimization problem by constructing a graph at a distance of $d_{max}$ from obstacles, or a lower distance between two obstacles. 
The path is found across this graph, so that the UAV travels either at $d_{max}$ from obstacles, through a ridge between two obstacles, or in a straight line between two obstacles.
The method does not scale well with the environment size as only a local map with a size of \vk{\numrange{3}{4} times the} sensor range is maintained and thus it does not allow global navigation in a larger environment.
In contrast to our method, which was verified in both outdoor and indoor/subterranean environment, the authors also only assume an outdoor environment.

For planning outside the local map boundaries, \cite{collins2020efficient} builds a sparse graph containing two types of vertices: the history of the robot's past positions to quickly navigate through an already explored environment, and the visibility nodes that represent potential pathways to new space.
The safety of local trajectories is ensured by including a collision probability penalization. 
This penalization is based on the pose estimate covariance in the cost function of the motion primitive selection.
The presented method lacks a technique for rerouting the sparse graph in a dynamic environment, where a previously free path can suddenly become obstructed.
The evaluation was also realized only in a simplified simulation and the method does not consider common real world issues such as imperfect sensor data, map drift, control error, and external disturbances.

\subsection{Contributions}
\label{section:contributions}
In context of the above discussed literature, the contributions of this work can be summarized as follows:
\begin{itemize}
  \item a novel dynamically built graph structure (\map{}) with clearance information that allows for rapid safety-aware path planning between any two points in known space and can handle \vk{arbitrarily structured and} changing environments,
  \item a method for quickly generating a lightweight volumetric graph representation (\segmap{}) of the environment, that can be shared with other robots using low-bandwidth communication and carry mission-critical information,
  \item experimental evaluation showing that planning on the \map{} can be orders of magnitude faster than state-of-the-art approaches for long-distance planning,
  \item validation of the proposed methods in real world scenarios, most importantly in the DARPA SubT competition, where robots were deployed into both real and simulated previously unvisited environments\vk{~\cite{petrlik2022uavs}}.
\end{itemize}

\section{Problem Description}
The planning problem tackled in this paper is the computation of long-distance paths through known space represented by an occupancy map in real time, while also weighing the path length and path safety.
The UAV is assumed to be modeled as a sphere with a radius $r_{min}$. 
The dynamic constraints of the UAV are not considered, as the obtained path is used as an input into a local planner.
Formally, we define the planning problem as finding a solution path $\mathbf{x}(k)$ parametrized by $k \in [0,1]$ which minimizes the criterion
\begin{equation}
  J(\mathbf{x}(k)) = L(\mathbf{x}(k))+ Z(\mathbf{x}(k)),
  \label{eq:optimization_criterion}
\end{equation}
divided into the path length $L(\mathbf{x}(k))$ and path risk $Z(\mathbf{x}(k))$ defined as
\begin{equation}
  L(\mathbf{x}(k)) = \int_0^{1} \left\Vert\mathbf{x}'(k)\right\Vert  dk
\end{equation}
\begin{equation}
  Z(\mathbf{x}(k)) = \xi \int_0^{1} \left\Vert\mathbf{x}'(k)\right\Vert \left[ \max(0, d_{max}- d(\mathbf{x}(k)))\right]^2 dk
\end{equation}
while satisfying
\begin{equation}
  \min_{k \in (0,1)} d(\mathbf{x}(k)) > r_{min}
\end{equation}
\vk{where $d(\mathbf{x})$ is the nearest obstacle distance at point $\mathbf{x}$ and $\xi$ is a parameter specifying the weight of the path safety on the overall cost of the path.
The cost cutoff distance $d_{max}$ is determined based on the UAV's size and the maximum possible deviation caused by disturbances and control errors.}

The second problem to be solved is the ability to efficiently generate a lightweight topological-volumetric representation of the visited part of the environment at any time, which can carry mission-relevant information about sections of the environment to share with the rest of the robot and human team.
We further specify the following requirements and assumptions for the problem:
\begin{enumerate}
  \item Mission: to explore the given environment and share exploration and mission-relevant information to other UAVs and human operators \vk{when communication is available}.
  \item Environment: unstructured and unknown dynamic subterranean or outdoor environment with narrow passages, sections with unreliable low bandwidth communication, and without access to GNSS.
  \item UAV Platform: assumed to carry depth sensors (LiDAR or depth cameras) and build an occupancy map of the environment onboard while localizing itself using a SLAM method.
    \vk{The UAV is modeled as a sphere with no physical collision tolerance of the platform, such as in \cite{de2021resilient}}.
  \item Planning time: the method must be able to find paths to goals which are hundreds of metres away. 
    This has to be done in tens of milliseconds, so that the UAV can evaluate paths to all potential goals at any time, as sometimes a safe, but distant goal might be preferable to a nearer goal that is reachable only by a risky passage.
\end{enumerate}


\section{The \map\ }
In this section, the proposed multi-layer graph structure, the method of incrementally building the structure, and the method for quickly extracting a lightweight topological-volumetric map from the structure are described.
\subsection{Topological map representation}
\label{section:map_representation}
We propose representing the topology of the environment as an undirected graph of spheres $\mathcal{G}$ divided into roughly convex segments $\mathcal{S}$.
Each node $\nu$ on the sphere graph is assigned a position $\mathbf{p}_\nu$, a radius $r_\nu$ equal to the nearest obstacle distance from $\mathbf{p}_\nu$, and a label $\sigma_\nu$ signifying which segment the node belongs to.
We connect every two spheres for which the radius of their intersection is bigger than $r_{min}$.
For planning, the graph of spheres, illustrated in~\autoref{fig:map_planning_illustration}, is similar to an ESDF used in \cite{voxblox} as it inherently encodes the distance from obstacles.
\vk{However, in contrast to \cite{voxblox}, the proposed map representation is not bound to a fixed resolution grid, where narrow passages can appear blocked due to discretization errors when the voxel size is set too high, and where the planning can be greatly slowed down if the voxels are too small.}
The graph of spheres is also stored in an octree structure for quick nearest-neighbor querying and short insertion and deletion time.

The graph of spheres is divided into roughly convex segments of nodes $\mathcal{S}$, which form a topological graph.
The convexity condition is not strict, since the paths between portals are collision-free intra-segment paths introduced in~\autoref{section:segmentation}, and not straight lines as in~\cite{topomap}.
We add a connection to the topological graph for every two regions which have some connected nodes, and store the pair of nodes $\nu_1 \in \sigma_1, \nu_2 \in \sigma_2$ with the largest sphere-sphere intersection radius.
We call these pairs \textit{portals} between segments, as in~\cite{topomap}.
To allow for efficient long-distance planning, we also iteratively compute and store optimal \vk{(according to~\eqref{eq:optimization_criterion})} paths between every two portals of a given segment, as illustrated in~\autoref{fig:map_planning_illustration}.

\subsection{Growing the \map\ }

\begin{figure}[ht]
  \definecolor{green}{HTML}{3FA63F}
  \definecolor{red}{HTML}{EB4E3F}
  \definecolor{blue}{HTML}{52B8E8}
  \newcommand{\greendot}{\raisebox{0pt}{\tikz{\draw[green,fill=green] (0,0) circle (2.0pt);}}}
  \newcommand{\reddot}{\raisebox{0pt}{\tikz{\draw[red,fill=red] (0,0) circle (2.0pt);}}}
  \newcommand{\bluedot}{\raisebox{0pt}{\tikz{\draw[blue,fill=blue] (0,0) circle (2.0pt);}}}
  \def\subfigwidth{0.49\linewidth}
  \centering
     \begin{subfigure}[b]{\subfigwidth}
      \begin{tikzpicture}
        \centering
        \node[anchor=south west,inner sep=0] (a) at (0,0) {\adjincludegraphics[width=1.0\linewidth, trim={{0.0\width} {0.0\height} {0.0\width} {0.0\height}}, clip=true]{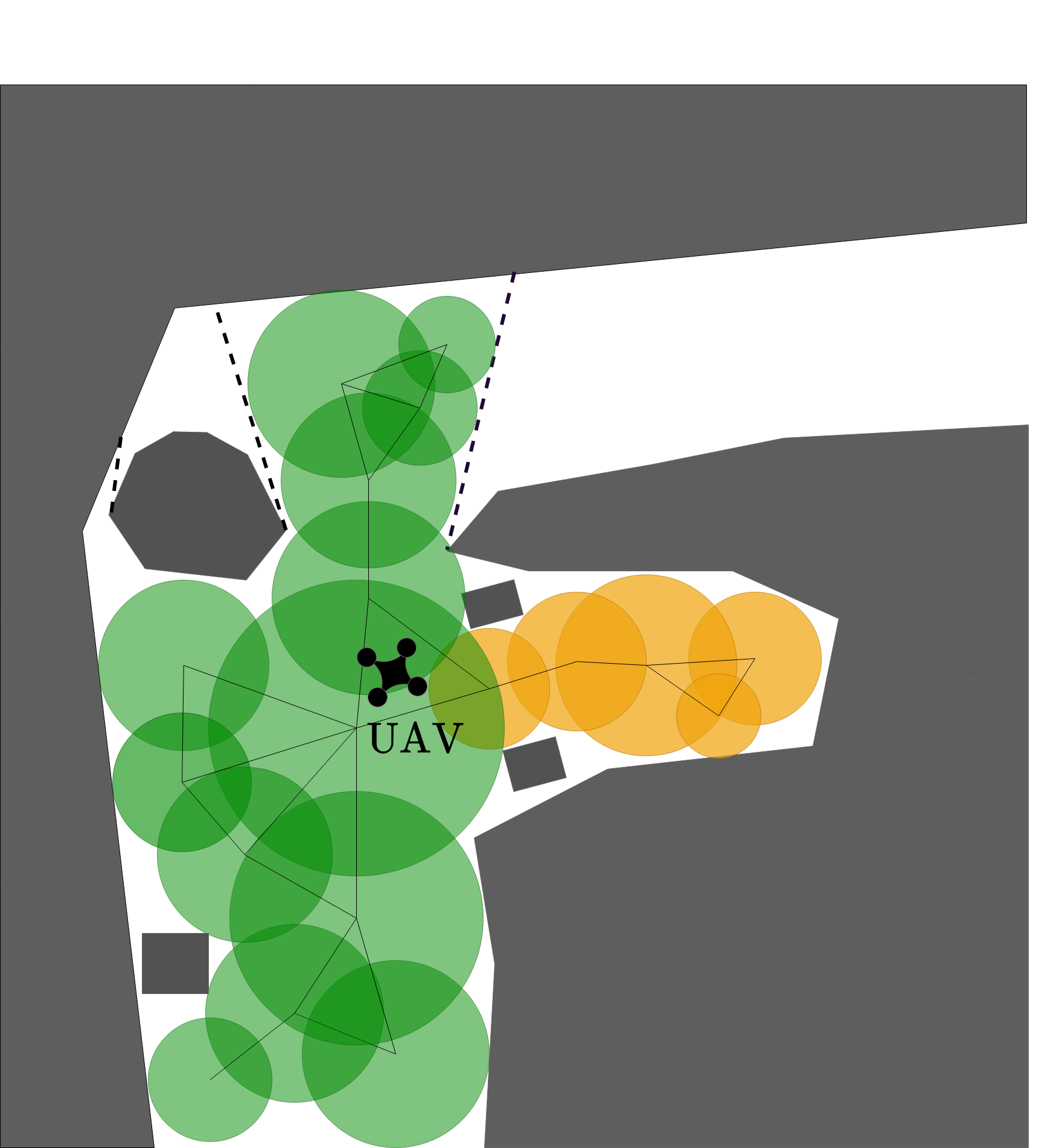}};
        \begin{scope}[x={(a.south east)},y={(a.north west)}]
          \node[align=center] at (0.9, 0.075) {\footnotesize \color{white}(a)};
        \end{scope}
      \end{tikzpicture}
     \end{subfigure}
     \hfill
     \begin{subfigure}[b]{\subfigwidth}
      \begin{tikzpicture}
        \centering
        \node[anchor=south west,inner sep=0] (a) at (0,0) {\adjincludegraphics[width=1.0\linewidth, trim={{0.0\width} {0.0\height} {0.0\width} {0.0\height}}, clip=true]{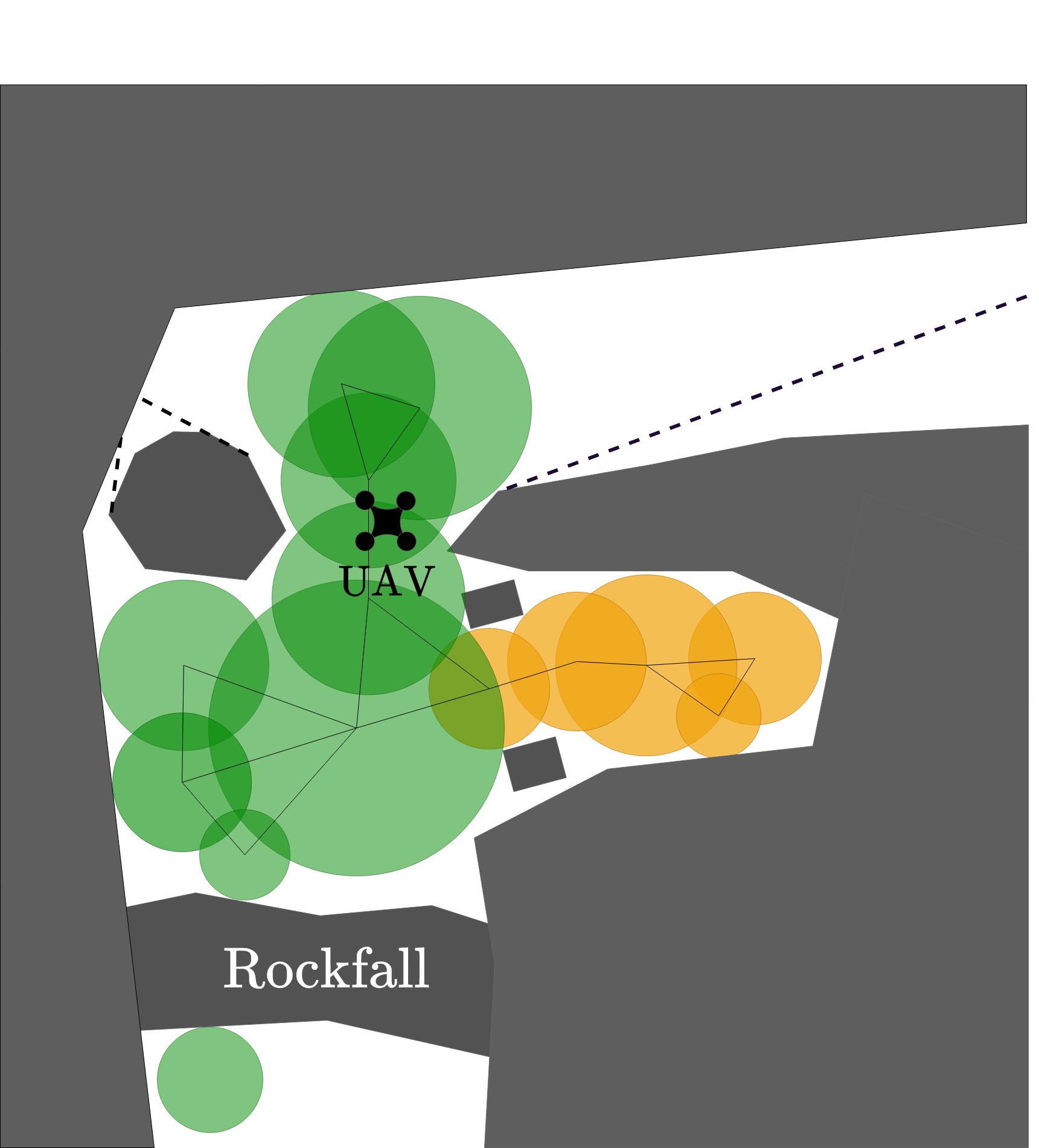}};
        \begin{scope}[x={(a.south east)},y={(a.north west)}]
          \node[align=center] at (0.9, 0.075) {\footnotesize \color{white}(b)};
        \end{scope}
      \end{tikzpicture}
     \end{subfigure}
     \begin{subfigure}[b]{\subfigwidth}
      \begin{tikzpicture}
        \centering
        \node[anchor=south west,inner sep=0] (a) at (0,0) {\adjincludegraphics[width=1.0\linewidth, trim={{0.0\width} {0.0\height} {0.0\width} {0.0\height}}, clip=true]{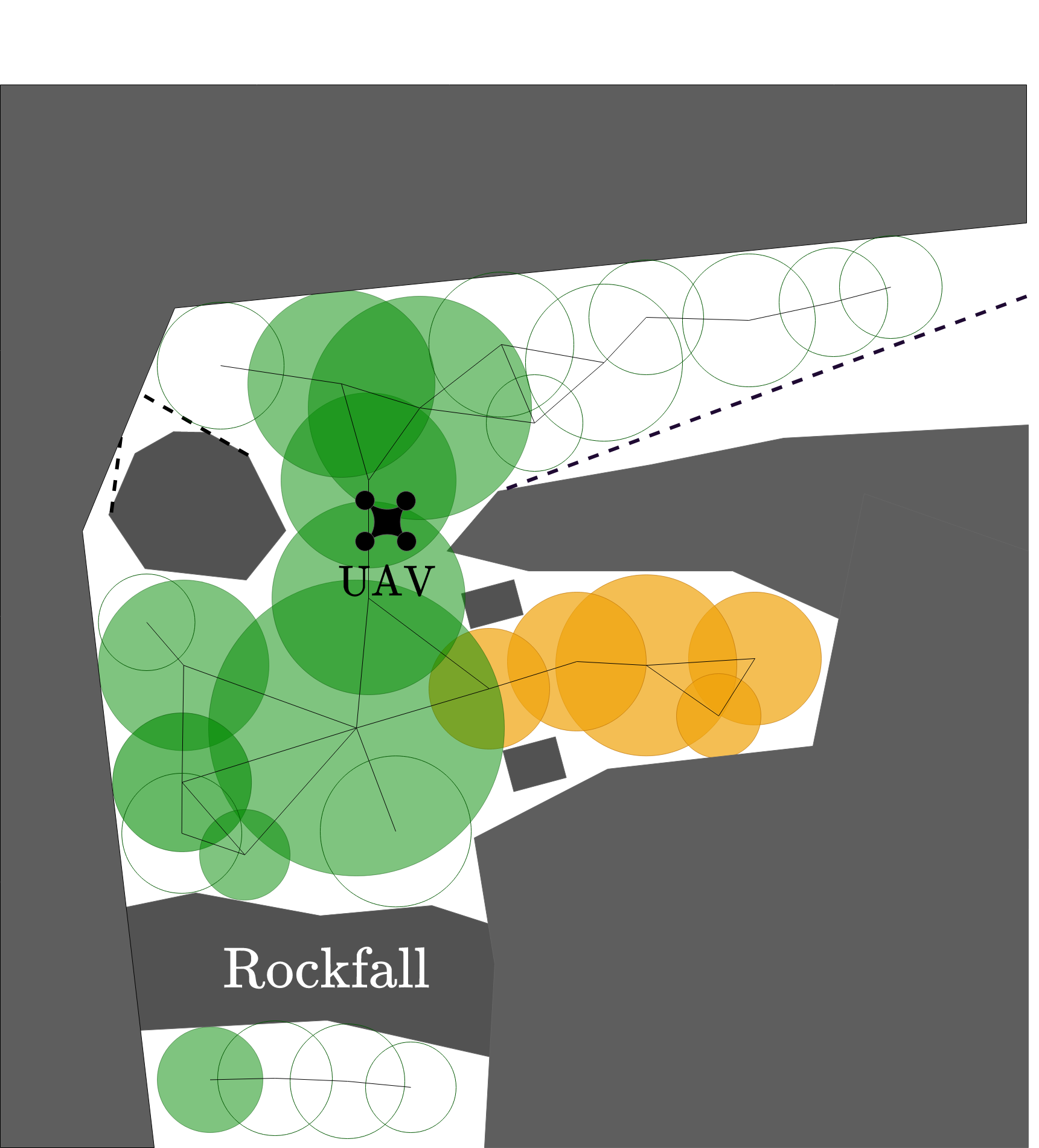}};
        \begin{scope}[x={(a.south east)},y={(a.north west)}]
          \node[align=center] at (0.9, 0.075) {\footnotesize \color{white}(c)};
        \end{scope}
      \end{tikzpicture}
     \end{subfigure}
     \hfill
     \begin{subfigure}[b]{\subfigwidth}
      \begin{tikzpicture}
        \centering
        \node[anchor=south west,inner sep=0] (a) at (0,0) {\adjincludegraphics[width=1.0\linewidth, trim={{0.0\width} {0.0\height} {0.0\width} {0.0\height}}, clip=true]{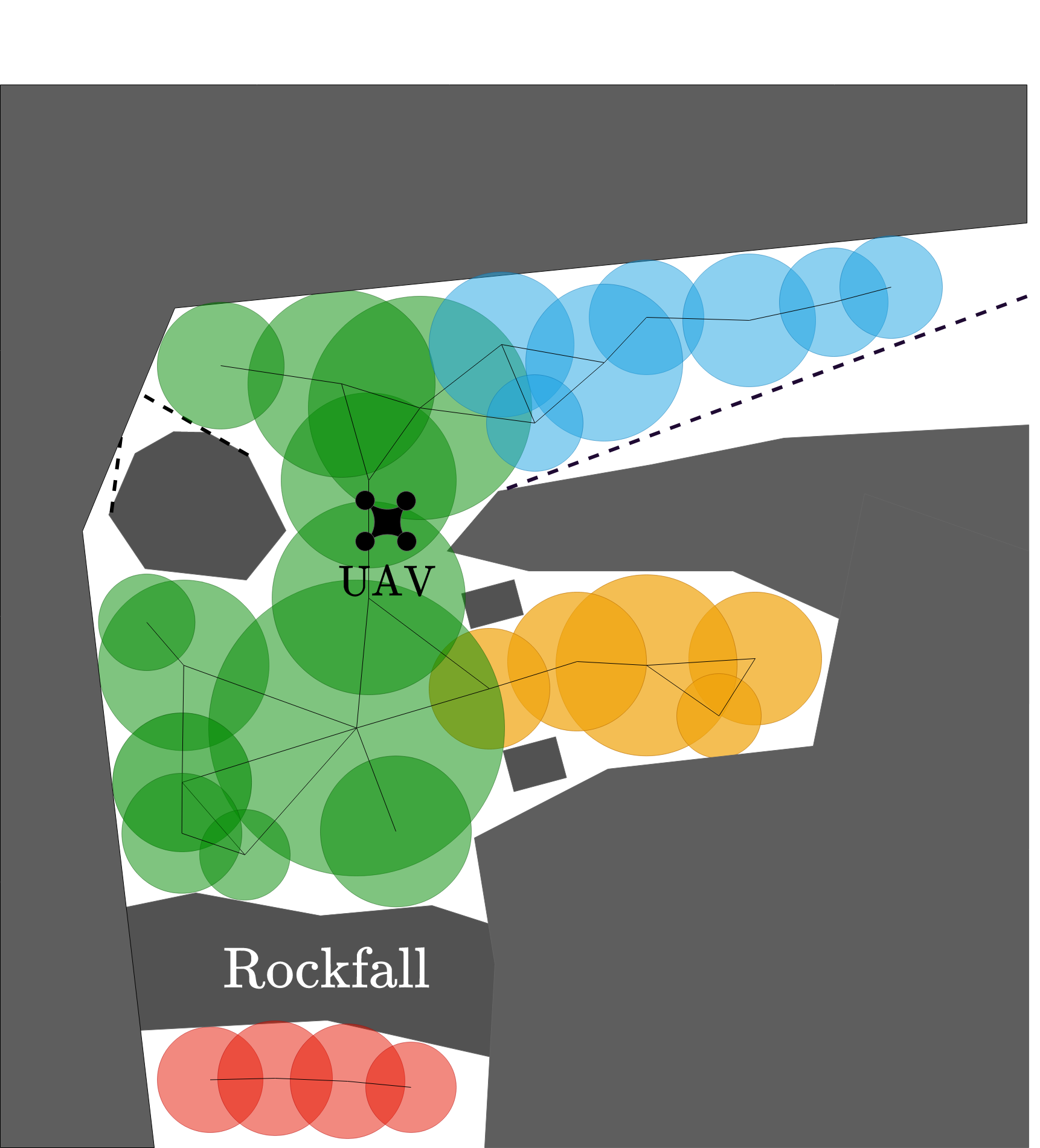}};
        \begin{scope}[x={(a.south east)},y={(a.north west)}]
          \node[align=center] at (0.9, 0.075) {\footnotesize \color{white}(d)};
        \end{scope}
      \end{tikzpicture}
     \end{subfigure}

  \caption{
    Outline of one iteration of the \map{} update algorithm. 
    The dashed lines represent the boundary of known free space and unknown space. 
    (a): Map at the end of the previous iteration.
    (b): Sphere radius update and pruning of unsafe and redundant nodes.
    (c): Sampling, creating and connecting new nodes into the graph.
    (d): Dividing disconnected (\protect\greendot) segment into two segments (\protect\greendot,\protect\reddot), expanding existing segments and growing a new segment (\protect\bluedot).
  }
  \label{fig:growing_illustration}
\end{figure}
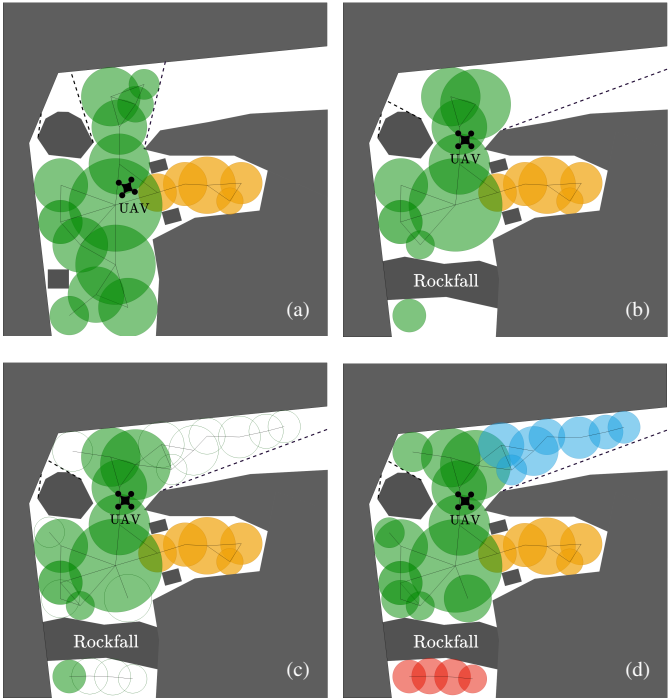

The proposed structure is built incrementally during a mission at a low rate of \SI{2}{\hertz}. 
The graph of spheres is grown so that the spheres cover the entire reachable known space of the environment, and  redundant nodes are pruned from the sphere graph to speed up computations.
During a single update iteration, only the part of the map inside a cube with a pre-defined size centered on the UAV is modified. 
As input, the update iteration only needs the current position of the UAV and \vk{the most recent} occupancy information about the space in the cube (meaning the information whether a given point in the cube is occupied, free, or unknown).
Any \vk{onboard-running occupancy mapping method} can be used to provide this information, such as the efficient occupancy map implementations OctoMap \cite{octomap} or UFOMap \cite{duberg2020ufomap}.

To enable fast querying of the nearest obstacle distance at any given point, we compute a k-D tree from occupied points (e.g. centroids of occupied voxels of an occupancy grid) in the cube of nearby space.
To prevent the spheres from expanding into unknown space, we also add points that lie at the boundary of free and unknown space into the k-D tree.
Finding these boundary points (referred to as frontiers) is non-trivial in most occupancy maps and usually slower than constructing the k-D tree, but it is necessary for safe path planning and also useful for exploration.

\vk{The \map\ is built in approximately constant-time iterations, which converge to an optimal distribution of spheres with increasing number of iterations in a given area.
As such, it can occasionally happen that the spheres do not cover the entire reachable space, which can cause some points to be unreachable through the \map\ or cause the found paths to be far from optimal.
However, this can be handled by mission control.
}
A single \map\ update iteration\vk{, further outlined in~\autoref{fig:growing_illustration},} consists of three steps in this order:
\subsubsection{Recomputing and pruning step}
\label{section:pruning}
Because the safe planning requires the most recent occupancy information, we first recompute the radius of every nearby node $\nu \in \mathcal{N}_{near}$ and update the connections of the nodes for which the radius has changed, as shown in~\autoref{alg:pruning}. 
The radius computation is done by querying the nearest point distance from the current obstacle k-D tree \vk{$K$} for each node in the updated part of the sphere graph.
Since we assume a dynamic environment, some nodes can become untraversible. This can occur due to obstacles moving in the environment or noisy sensor data. 
These nodes are pruned from the graph.

Furthermore, some nodes can grow in size due to more free space being uncovered, making other nodes obsolete.
We mark a node as redundant if there exists a larger node, which covers at least a certain percentage of the node's volume.
All redundant nodes are then pruned.

\begin{algorithm}
  \footnotesize
  \captionsetup{labelformat=empty}
  \caption{\footnotesize{\textbf{Algorithm 1:} \map{} radius update and pruning step. Only nodes $\mathcal{N}_{near}$ in a bounding box around the UAV are updated.}}
  \label{alg:pruning}

  \algdef{SE}[SUBALG]{Indent}{EndIndent}{}{\algorithmicend\ }%
  \algtext*{Indent}
  \algtext*{EndIndent}

  \algnewcommand\graph{\mathcal{G}}
  \algnewcommand\nearnodes{\mathcal{N}}

  \algnewcommand\algorithmicinput{\textbf{Input:~}}
  \algnewcommand\Input{\State\algorithmicinput}%
  \algrenewcommand\alglinenumber[1]{\scriptsize #1:}
  \begin{algorithmic}[1]
  \scriptsize

  \Procedure{recompute\map{}}{}

  \Input

  \Indent

  \State $\graph$\Comment{\map{} graph}
  \State $K$\Comment{obstacle k-D tree}
    \State $r_{min}$\Comment{minimal distance from obstacles}

  \EndIndent

    \For{$\nu \in \nearnodes_{near}$} \Comment{nodes to be updated in a cube around UAV}
    \State $r_{\nu} \gets \function{nearestObstacleDistance($\mathbf{p}_\nu, K$)}$
  \If{$r_\nu < r_{min}$ }
    \State $\graph \gets \graph \setminus \nu$
  \Else
    \State $\function{updateNodeConnections($\nu, \graph$)}$
\EndIf
  \EndFor
  \For{$\nu \in \nearnodes_{near}$}
    \If{$\function{isRedundant($\nu, \graph$)}$}
      \State $\graph \gets \graph \setminus \nu$
\EndIf
  \EndFor
\EndProcedure

\end{algorithmic}

\end{algorithm}

\subsubsection{Expansion step}
In the next step, shown in~ \autoref{alg:expansion}, new non-redundant spheres are added to the graph by sampling points in the vicinity of the UAV.
\vk{The number of sampled points $S$ heavily influences the runtime of this step.}
Sampling a large amount of points will ensure the reachable space is filled with spheres, but it demands more computational resources. 
A sparser sampling method will speed up the expansion step, but \vk{in the newly seen large areas, it} might cause some reachable space to not be covered by the sphere graph.
In our implementation, we chose to sample one point per each voxel of the used occupancy octree, and also cast rays in random directions from the UAV and sample points along the rays.

\begin{algorithm}

  \footnotesize
  \captionsetup{labelformat=empty}
  \caption{\footnotesize{\textbf{Algorithm 2:} \map{} expansion step. New nodes that are not inside already existing nodes are added to the graph at sampled positions.}}
  \label{alg:expansion}

  \algdef{SE}[SUBALG]{Indent}{EndIndent}{}{\algorithmicend\ }%
  \algtext*{Indent}
  \algtext*{EndIndent}

  \algnewcommand\graph{\mathcal{G}}
  \algnewcommand\algorithmicinput{\textbf{Input:~}}
  \algnewcommand\Input{\State\algorithmicinput}%
  \algrenewcommand\alglinenumber[1]{\scriptsize #1:}
  \begin{algorithmic}[1]
  \scriptsize

  \Procedure{expand\map{}}{}

  \Input

  \Indent

    \State $\graph$\Comment{\map{} graph}
  \State $K$\Comment{obstacle k-D tree}

  \EndIndent

    \State $\mathcal{P}_s \gets \function{samplePointsInNearFreeSpace($\mathbf{p}_{uav}$)}$
    \For{$\mathbf{p}_s \in \mathcal{P}_s$}
    \State $\nu_{new} \gets \function{initNode()}$
    \State $\mathbf{p}_{\nu_{new}} \gets \mathbf{p}_s$
    \State $r_{\nu_{new}} \gets \function{nearestObstacleDistance($\mathbf{p}_s, K$)}$ 
    \State $\sigma_{\nu_{new}} \gets \emptyset$
    \If{$r_{\nu_{new}} \geq r_{min}$ \textbf{and not} $\function{isRedundant($\nu_{new}, \graph$)}$}
    \State $\graph \gets \graph \cup \nu_{new}$
    \State $\function{updateNodeConnections($\nu_{new}, \graph$)}$

  \For{$\nu' \in \function{connectedNodes($\nu_{new}, \graph$)}$}
  \If{$\function{isRedundant($\nu', \graph$)}$}
    \State $\graph \gets \graph \setminus \nu'$
\EndIf
  \EndFor
\EndIf
  \EndFor
  \EndProcedure
\end{algorithmic}
\end{algorithm}

\subsubsection{Segmentation step}
\label{section:segmentation}
In the last step, shown in~\autoref{alg:segmentation}, we update the segmentation of the map.
Each node $\nu$ in the graph, except for the nodes that were created in the current expansion step, belongs to some region $\sigma_{\nu}$.
In this step, we first assign each new node to a segment.
This is done by first expanding the existing regions using a flood fill over the sphere graph, limiting the maximum bounding sphere of the region to a radius $r_{exp}$.
Then, we iterate over the remaining newly added nodes without an assigned segment, and initialize and grow new segments (also bounded by $r_{exp}$) from them, until all nodes belong to some segment.

Since this approach can create a very granular segmentation of the graph, we attempt to merge all the regions that have been modified in this step with their neighboring regions.
A merge is allowed only if the resulting region does not exceed the bounding sphere radius $r_{merge}$, and also if the centers of the segments are mutually visible.
By using these conditions and setting $r_{exp}$ very low compared to $r_{merge}$, we can achieve roughly convex segmentation with low computational effort.

Additionally, we recompute and store the optimal paths $\mathcal{C}_{(p_1,p_2)}$ with respect to the criterion~\eqref{eq:optimization_criterion} between each pair of portals $(p_1, p_2)$ inside every segment that has been modified in the iteration.
A segment can become modified either by the segmentation step or by the update and pruning step, meaning that some of its nodes were altered or some edges between nodes were deleted.
These paths{\vk{, henceforth called intra-segment paths,}} are computed using the A* algorithm with the path cost and heuristic functions described in~\autoref{section:path_planning}, and then cached for rapid on-demand long-distance path planning.

\begin{algorithm}

  \footnotesize
  \captionsetup{labelformat=empty}
  \caption{\footnotesize{\textbf{Algorithm 3:} \map{} segmentation step. Intra-segment paths between portals are stored in cache $\mathcal{C}$ for faster planning.}}
  \label{alg:segmentation}

  \algdef{SE}[SUBALG]{Indent}{EndIndent}{}{\algorithmicend\ }%
  \algtext*{Indent}
  \algtext*{EndIndent}

  \algnewcommand\graph{\mathcal{G}}
  \algnewcommand\segment{\mathcal{S}}
  \algnewcommand\nearnodes{\mathcal{N}}
  \algnewcommand\cache{\mathcal{C}}
  \algnewcommand\algorithmicinput{\textbf{Input:~}}
  \algnewcommand\Input{\State\algorithmicinput}%
  \algrenewcommand\alglinenumber[1]{\scriptsize #1:}
  \begin{algorithmic}[1]
  \scriptsize

  \Procedure{segment\map{}}{}
  \Input

  \Indent

  \State $\segment_\graph$\Comment{\map{} graph segments}
    \State $r_{exp}$\Comment{maximum radius of segment expansion bounding sphere}
    \State $r_{merge}$\Comment{maximum radius of segment merging bounding sphere}

  \EndIndent

  \State $\segment_{near} \gets \function{getNearSegments($\segment_\graph$)}$
  \State $\segment_{near} \gets \function{splitDisconnectedSegments($\segment_{near}$)}$
  \For{$\sigma \in \segment_{near}$} \Comment{try expanding existing segments}
    \State $\sigma \gets \function{expandSegment($\sigma, r_{exp}$)}$
  \State $\function{recomputeSegmentPortals($\sigma$)}$
  \EndFor

    \For{$\nu \in \nearnodes_{near} , \sigma_{\nu} = \emptyset$} \Comment{grow new segments}
  \State $\sigma_{new} \gets \function{initSegment($\nu$)}$
  \State $\sigma_{new} \gets \function{expandSegment($\sigma_{new}$)}$
  \State $\segment_{near} \gets \segment_{near} \cup \sigma_{new}$
  \State $\function{recomputeSegmentPortals($\sigma_{new}$)}$
  \EndFor

    \For{$(\sigma_1, \sigma_2) \in \function{getAdjacentSegmentPairs($\segment_{near}$)}$} \Comment{merge segments}
    \If{$\function{canMergeSegments($\sigma_1, \sigma_2, r_{merge}$)}$}
    \State $\sigma_1 \gets \function{mergeSegments($\sigma_1, \sigma_2$)}$
  \State $\segment_{near} \gets \segment_{near} \setminus \sigma_2 $
  \EndIf
  \EndFor

  \For{$\sigma \in \segment_{near}$} \Comment{recompute intra-segment paths}
  \If{$\function{segmentWasAltered($\sigma$)}$}
  \For {$(p_1, p_2) \in \function{getSegmentPortals($\sigma$)}$}
  \State $\cache_{(p_1, p_2)} \gets \function{computePathInSegment($\sigma, p_1, p_2$)}$
  \EndFor
  \EndIf
  \EndFor
  \EndProcedure
\end{algorithmic}
\end{algorithm}

\subsection{Lightweight map creation and sharing}

Another key feature of segmentation of the sphere graph is that the roughly convex regions of space can be represented by convex 3D shapes (in our implementation, we used 4DOF bounding boxes{\vk{ determined by width, height, depth, and z axis rotation}, as shown in~\autoref{fig:map_compression_octomap}).
After each update iteration of the \map{}, we compute and store the best-fitting convex shape envelope{\vk{ of a segment's spheres} } for any nearby segments that were modified.
Therefore a topological-volumetric graph representation of the environment (\segmap{}) can be extracted at any time. 
Each vertex of the \segmap{} graph represents the pose and dimensions of a segment's envelope, with edges added to the graph for every two connected segments. 

This representation is more lightweight than both the \map{} itself and the underlying occupancy map, and can therefore be used in low-bandwidth situations, in which sharing the full information is not possible.
For \vk{SAR} missions, additional information (such as the percentage of surfaces covered by the UAV's cameras or information about frontiers in a given region) can easily be appended to each region. 
This information can then be shared with other robots during the mission for cooperation purposes, and also with human operators to whom this map representation can provide a good overview of the environment and the mission progress\vk{~\cite{petrlik2022uavs}}. 

\label{section:map_sharing}

\vk{
\subsection{Computational complexity}
The leading factors in every update iteration's complexity include the determination of the sphere's radius around a given point as $\mathcal{O}(\log |K|)$, and checking the nearest spheres for possible intersections, which is $\mathcal{O}(n_c \log |\mathcal{N}_{near}|)$. Herein, $n_c$ is the average number of nearest spheres that need to be checked, depending on the density of spheres and the largest sphere in the update bounding box.
Hence, the entire update is $\mathcal{O}( (|\mathcal{N}_{near}| + |S|) (n_c \log |\mathcal{N}_{near}| + \log|K|))$.
By only updating in a small area around the UAV, the update time does not grow with the total number of spheres in the \map, rather it depends on the sphere density in the local area.
By pruning the sphere graph, the method keeps $n_c$ and $|\mathcal{N}_{near}|$ reasonably low, thus also keeping the runtime low.
}

\section{Planning in the \map{}}
\label{section:path_planning}
For planning paths on a graph or grid with the optimization criterion~\eqref{eq:optimization_criterion}, the criterion first needs to be discretized.
Let us assume a segment of a path between two nearby points $\mathbf{p}_1, \mathbf{p}_2$ that have the distance from the nearest obstacle equal to $r_{\mathbf{p}_1}, r_{\mathbf{p}_2}$ respectively.
Naturally, the length increment can be written as 
\begin{equation}
 \Delta L_{\mathbf{p}_1 \mathbf{p}_2} = \left\Vert \mathbf{p}_1 - \mathbf{p}_2 \right\Vert,
\end{equation}
but the computation of the safety increment $\Delta Z$ is trickier if we only access obstacle distances from a k-D tree.
The integral is approximated by defining the increment as
\begin{equation}
  \Delta Z_{\mathbf{p}_1 \mathbf{p}_2} =\xi  \mkern-4mu \left [\max \mkern-4mu \left(0, d_{max} - \frac{r_{\mathbf{p}_1} + r_{\mathbf{p}_2}}{2} \right) \right]^2 \mkern-14mu \left\Vert\mathbf{p}_1-\mathbf{p}_2\right\Vert.
\end{equation}
During planning, we also ensure that there are no points along the line between the points $\mathbf{p}_1$ and $\mathbf{p}_2$ (using the size of the intersection of spheres around $\mathbf{p}_1$ and $\mathbf{p}_2$ with radii $r_{\mathbf{p}_1}$ and $r_{\mathbf{p}_2}$) that would be closer to obstacles than the minimal allowed distance $r_{min}$. 
We use A* with a transition cost between two points given by 
\begin{equation}
  g(\mathbf{p}_1, \mathbf{p}_2) = \Delta L_{\mathbf{p}_1 \mathbf{p}_2} + \Delta Z_{\mathbf{p}_1 \mathbf{p}_2}
\label{eq:criterion_discretization}
\end{equation}
and a heuristic
\begin{equation}
  h(\mathbf{p}_2) = \left\Vert \mathbf{p}_2 - \mathbf{p}_{goal} \right\Vert
\end{equation}
for planning in the \map{}.

The main advantage of the segmentation and caching of the optimal paths between portals is that whenever a path between points A and B is queried, we only need to compute paths across the sphere graph from the start and end points to the portals of the segments the points are in.
The rest of the path can be planned using the cached intra-segment paths, as shown in~\autoref{fig:map_planning_illustration}.

This technique provides a major speedup of 2 orders of magnitude over computing the path on the entire sphere graph, as shown in~\autoref{tab:planning_results_multipath}, but the resulting paths are forced to pass through the portals, through which the optimal path might not pass.

\begin{figure}[H]
  \newcommand{\blueline}{\raisebox{2pt}{\tikz{\draw[blue,solid,line width = 1.0pt](0,0) -- (5mm,0);}}}
  \newcommand{\blackline}{\raisebox{2pt}{\tikz{\draw[black,solid,line width = 1.0pt](0,0) -- (5mm,0);}}}
  \centering
  \begin{tikzpicture}
    \centering
    \node[anchor=south west,inner sep=0] (a) at (0,0) {\adjincludegraphics[width=1.0\linewidth, trim={{0.0\width} {0.0\height} {0.0\width} {0.0\height}}, clip=true]{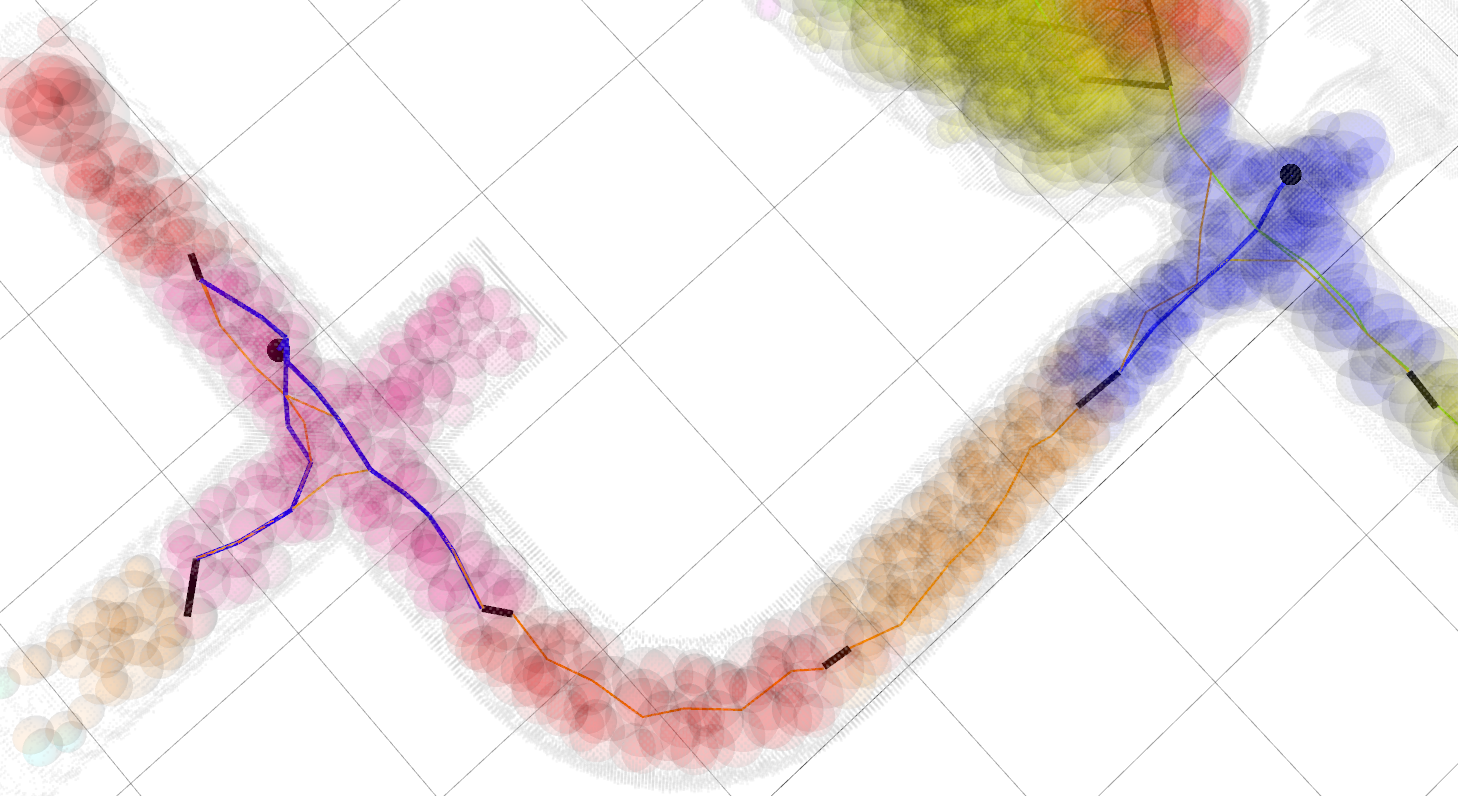}};
    \begin{scope}[x={(a.south east)},y={(a.north west)}]
      \node[align=center] at (0.214, 0.60) {\scriptsize \color{black}A};
      \draw[black,fill=black] (0.190,0.56) circle (2.0pt);
      \node[align=center] at (0.913, 0.815) {\scriptsize \color{black}B};
      \draw[black,fill=black] (0.885,0.78) circle (2.0pt);
      \draw [latex-latex](0.845,0.125) -- (0.945,0.305);
      \node[align=center] at (0.93, 0.175) {\scriptsize \color{black}\SI{10}{\meter}};
    \end{scope}
  \end{tikzpicture}
  \caption{Illustration of the proposed planning approach. When a path from A to B is queried, only a few paths (\protect\blueline) need to be computed across the sphere graph, namely the paths leading from A to the portals \vkcaption{(\protect\blackline)} of the segment that A is in, and then paths from portals of B's segment to B, when the portals are reached. The cached intra-segment paths{\vkcaption{ (thin colored lines) between portals}} are used for the rest of the computation.}
  \label{fig:map_planning_illustration}
\end{figure}
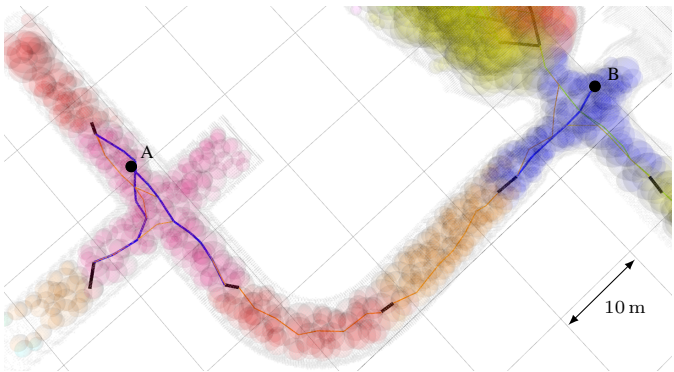


\section{Experiments}
\label{section:experiments}
This section aims to validate the proposed method on multiple simulated environments and compare \map\ with state-of-the-art methods. 
Furthermore, we analyze the runtime of the \map\ growing steps and show that the lightweight map generated from the \map\ is substantially smaller than a binary occupancy octree even with a lowered resolution, while keeping details about the environment connectivity and the size and geometric shape of each region.
We also demonstrate deployment of the proposed method onboard a real UAV platform in both subterranean and outdoor environment.

\subsection{Planning analysis}
To demonstrate the ability of the \map\ to quickly generate paths optimizing the criterion~\eqref{eq:optimization_criterion}, we conducted two experiments in simulated environments where we compare the proposed planning approach with the following state-of-the-art planning methods:
\begin{itemize}
\item 
\textit{A* through an occupancy grid:} A simple solution for path planning is to run A* in an occupancy grid with a fixed voxel size.
Using the transition cost and heuristic defined in~\autoref{section:path_planning} and a granular voxel size, this approach can find paths that are closer to the optimum defined in~\eqref{eq:optimization_criterion} than the \map{} approach, but as we show in \autoref{tab:planning_results_multipath}, such a method is unusable on large distances due to the long computation times.
\item 
\textit{RRT* through an occupancy grid:} Additionally, we compare the proposed planning approach with a popular planning algorithm, the asymptotically optimal RRT* \cite{karaman2011sampling}.
The path used for the comparison is the initial solution, i.e., the path is not further optimized by continuing the tree growth after finding the solution, which would generate solutions closer to the optimum, at the cost of increased computation time.
\end{itemize}

Firstly, we compared the planning methods in a situation where the UAV needs to find paths to a large amount of goals that can be up to \SI{300}{\meter} away from the UAV.
We used a standard laptop computer, a \SI{0.2}{\meter} resolution occupancy octree for the \map{} building and a \SI{0.8}{\meter} resolution for the grid-based A* computation.
The safety-based planning parameters were set to $\xi = 7, d_{max}=\SI{2}{\meter}, r_{min}=\SI{0.8}{\meter}$.
Results of this comparison are presented in~\autoref{tab:planning_results_multipath}, and found paths are shown in~\autoref{fig:planning_visualization_multipath}~(a).
It is evident from the results, that the state-of-the-art methods are too slow for such situations and do not meet the requirements set by this \paper{}.

In the second planning experiment, we analyzed the path length and risk as defined in~\eqref{eq:optimization_criterion} for paths found by the compared methods to a single goal.
To better outline the differences between the approaches, a \SI{0.4}{\meter} resolution was used for the grid-based A*. 
The same safety-based planning parameters as in the first experiment were used. 
The results of this analysis can be seen in~\autoref{fig:planning_visualization_multipath}~(b) and~\autoref{tab:planning_results}.
As can be seen, our approach outperforms the state-of-the art methods by orders of magnitude, while still being close to the optimal solutions found by grid-based A*.

\begin{figure}[h!]
  \definecolor{cyan}{HTML}{01FFFF}
  \newcommand{\blackdot}{\raisebox{0pt}{\tikz{\draw[black,fill=black] (0,0) circle (2.0pt);}}}
  \newcommand{\greenline}{\raisebox{2pt}{\tikz{\draw[green,solid,line width = 1.0pt](0,0) -- (5mm,0);}}}
  \newcommand{\cyanline}{\raisebox{2pt}{\tikz{\draw[cyan,solid,line width = 1.0pt](0,0) -- (5mm,0);}}}
  \newcommand{\redline}{\raisebox{2pt}{\tikz{\draw[red,solid,line width = 1.0pt](0,0) -- (5mm,0);}}}
  \newcommand{\orangeline}{\raisebox{2pt}{\tikz{\draw[orange,solid,line width = 1.0pt](0,0) -- (5mm,0);}}}
  \newcommand{\blueline}{\raisebox{2pt}{\tikz{\draw[blue,solid,line width = 1.0pt](0,0) -- (5mm,0);}}}
  \centering
  \begin{tikzpicture}
    \centering
    \node[anchor=south west,inner sep=0] (a) at (0,0) {\adjincludegraphics[width=1.0\linewidth, trim={{0.0\width} {0.0\height} {0.0\width} {0.0\height}}, clip=true]{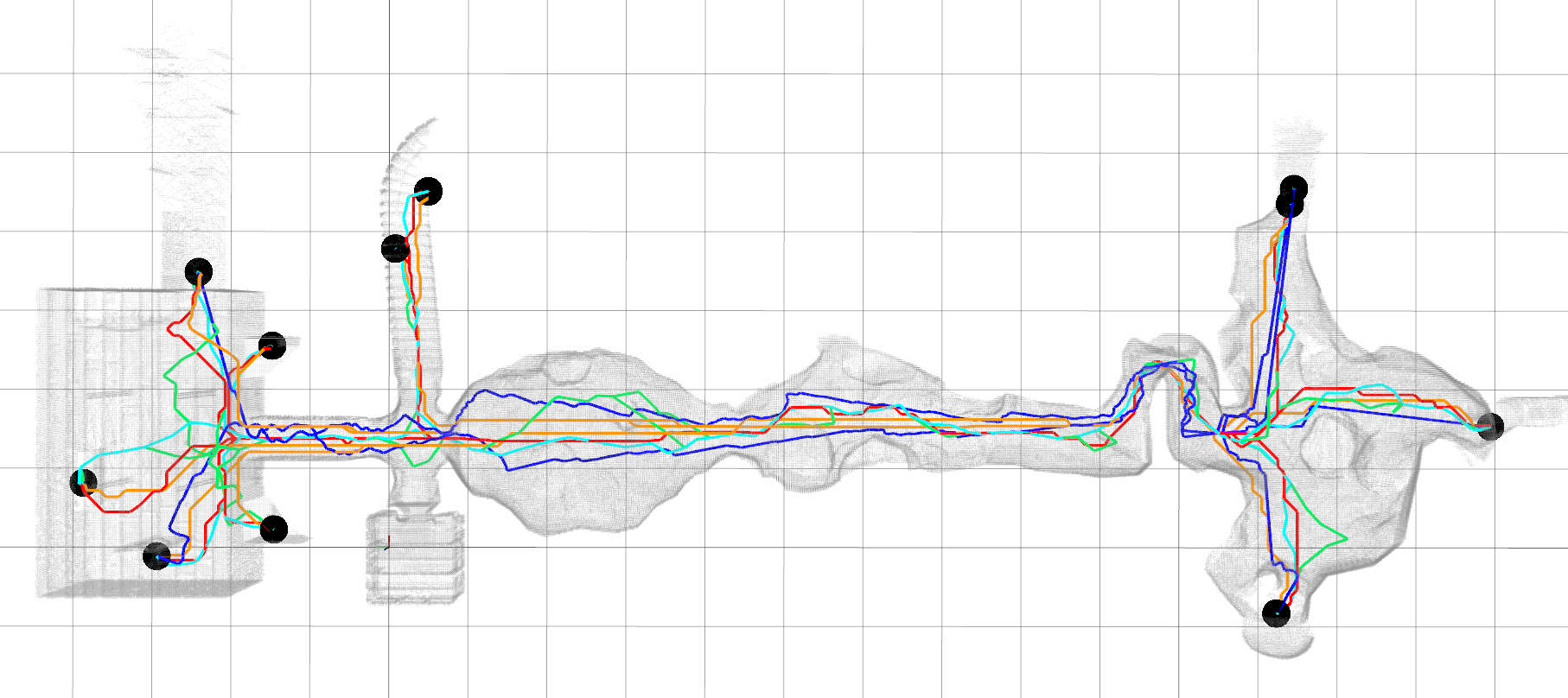}};
    \begin{scope}[x={(a.south east)},y={(a.north west)}]
      \node[align=center] at (0.755, 0.30) {\scriptsize \color{black}start};
      \node[align=center] at (0.05, 0.92) {\footnotesize \color{black}(a)};
      \draw[black,fill=black] (0.78,0.38) circle (3.0pt);
      \draw [latex-latex](0.903,0.12) -- (0.952,0.12);
      \node[align=center] at (0.94, 0.07) {\scriptsize \color{black}\SI{10}{\meter}};
    \end{scope}
  \end{tikzpicture}%
  \vspace{0.5em}
  \begin{tikzpicture}
    \centering
    \node[anchor=south west,inner sep=0] (a) at (0,0) {\adjincludegraphics[width=1.0\linewidth, trim={{0.0\width} {0.0\height} {0.0\width} {0.0\height}}, clip=true]{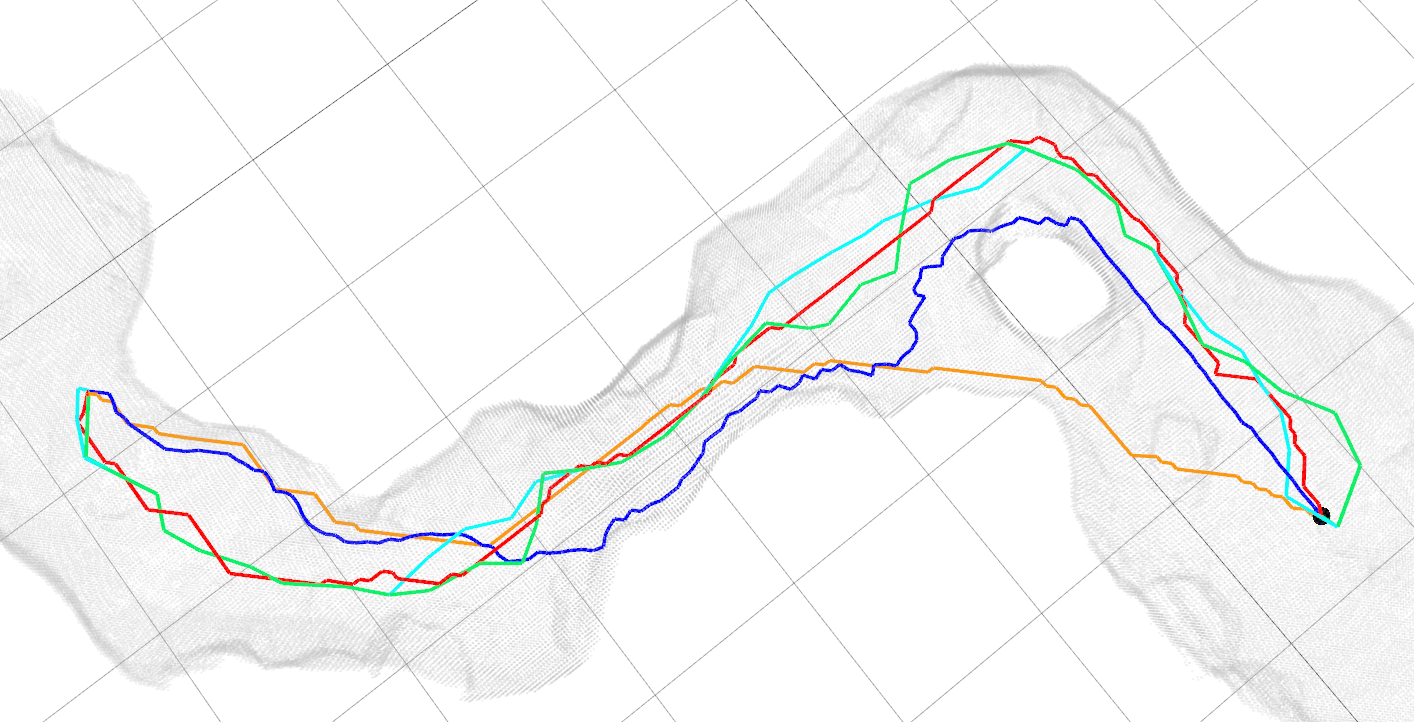}};
    \begin{scope}[x={(a.south east)},y={(a.north west)}]
      \node[align=center] at (0.05, 0.92) {\footnotesize \color{black}(b)};
      \node[align=center] at (0.09, 0.52) {\scriptsize \color{black}start};
      \draw[black,fill=black] (0.06,0.47) circle (2.0pt);
      \node[align=center] at (0.96, 0.245) {\scriptsize \color{black}goal};
      \draw[black,fill=black] (0.935,0.285) circle (2.0pt);
      \draw [latex-latex](0.86,0.95) -- (0.935,0.79);
      \node[align=center] at (0.94, 0.90) {\scriptsize \color{black}\SI{10}{\meter}};
    \end{scope}
  \end{tikzpicture}
  \caption{
    Visualization of path planning to multiple potential exploration goals (a) and to a single goal (b) using the following methods: 
    path generated by A* on the \map\ using cached intra-segment paths (\protect\greenline);
    the path planned through the entire sphere graph (\protect\cyanline);
    grid-based A* using cost function \eqref{eq:optimization_criterion} (\protect\redline); 
    grid-based A* with path cost equal to path length on the occupancy grid with voxel resolution $r=\SI{0.8}{\meter}$ (a) and $r=\SI{0.4}{\meter}$ (b) (\protect\orangeline);
    RRT* path (\protect\blueline).
  }
  \label{fig:planning_visualization_multipath}
\end{figure}
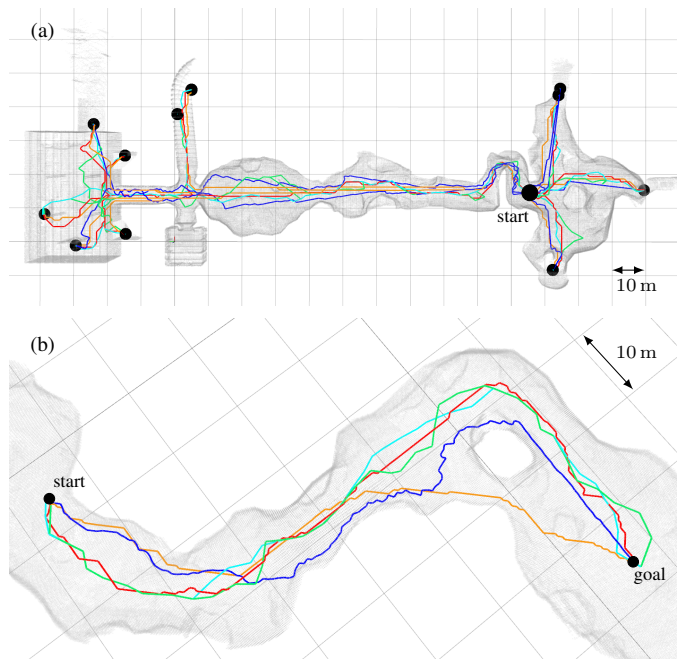

\begin{table}[h!]
  \centering
\setlength\tabcolsep{0.7em}
\begin{tabular}{llll}
\toprule
  \textbf{Planner} & \textbf{Time [ms]} & \textbf{Found paths}  &\textbf{Mean Cost [-]}\\
\midrule
  Grid A* & 59524 & 11/11 & 270.36\\
  Grid A* length-only & 58795 & 11/11 & 477.39 \\
  RRT* & 96830 & 6/11 & 385.05 \\
  \map\ & 1868 & 11/11 & \textbf{257.77} \\
  \map\ caching & \textbf{22} &  11/11 & 322.73\\
\bottomrule
\end{tabular}
  \caption{
    Quantitative results for the multiple-goal path planning experiment in a simulated environment, shown in~\autoref{fig:planning_visualization_multipath}~(a).
    The cost of each path was computed by summing the discretized transition cost~\eqref{eq:criterion_discretization}. 
    The computation time is summed for all found paths.
    Even with a \SI{10}{\second} planning timeout RRT* did not manage to find 5 of the paths.
    Compared to the state-of-the-art methods which find the paths in tens of seconds, planning over the entire sphere graph of the \map{} runs for only \SI{1.8}{\second}, and if the intra-segment path caching is used, the paths are found in only \SI{22}{\milli\second}.
}
  \label{tab:planning_results_multipath}
\end{table}

\begin{table}[h!]
  \centering
\setlength\tabcolsep{0.6em}
\begin{tabular}{lllll}
\toprule
  \textbf{Planner} & \textbf{Time [ms]} & \textbf{Length [m]} & \textbf{Risk [-]} &\textbf{Cost [-]}\\
\midrule
  Grid A* & 28064 & 101.11 & 33.88 & 134.99\\
  Grid A* length-only & 15376 & 80.33 & 376.21 & 456.54\\
  RRT* & 544 & 96.79 & 413.61 & 510.40\\
  \map\ & 56 & \textbf{104.14} & \textbf{45.98} & \textbf{150.12}\\
  \map\ caching  & \textbf{0.496} & 110.46 & 49.37 & 159.83\\
\bottomrule
\end{tabular}
  \caption{
    Quantitative results for the single-goal planning experiment in a simulated cave environment, shown in~\autoref{fig:planning_visualization_multipath}~(b).
    The cost of each path was computed by summing the discretized transition cost~\eqref{eq:criterion_discretization}.
  Our method is 3 orders of magnitude faster than the initial solution found by RRT* and 5 orders of magnitude faster than grid-based A* with a $\SI{0.4}{\meter}$ resolution.
  The \vkcaption{A* optimization of the path length only} naturally finds the shortest path, but as shown in~\autoref{fig:planning_visualization_multipath}~(b), the path leads through the riskier narrow passage.
}
  \label{tab:planning_results}
\end{table}


\subsection{Iteration runtime analysis}
To demonstrate the light computational load of building the \map{}, we conducted an experiment in a simulated exploration mission through a vast cave environment identical in scale to the ones shown in the other experiments, and measured the runtime of individual parts of the \map\ growing process. 

The results of this experiment can be seen in~\autoref{fig:spheremap_runtimes}.
The experiment demonstrates that the update iteration of the \map{} reaches a mean computation time of approx. \SI{150}{\milli\second}.
It is important to note that the overall growth runtime depends heavily on the expansion step and can be sped up by using a sparser sampling method.

\begin{figure}[thpb]
  \centering
  \adjincludegraphics[width=0.8\linewidth, trim={{0.00\width} {0.06\height} {0.00\width} {0.05\height}}, clip]{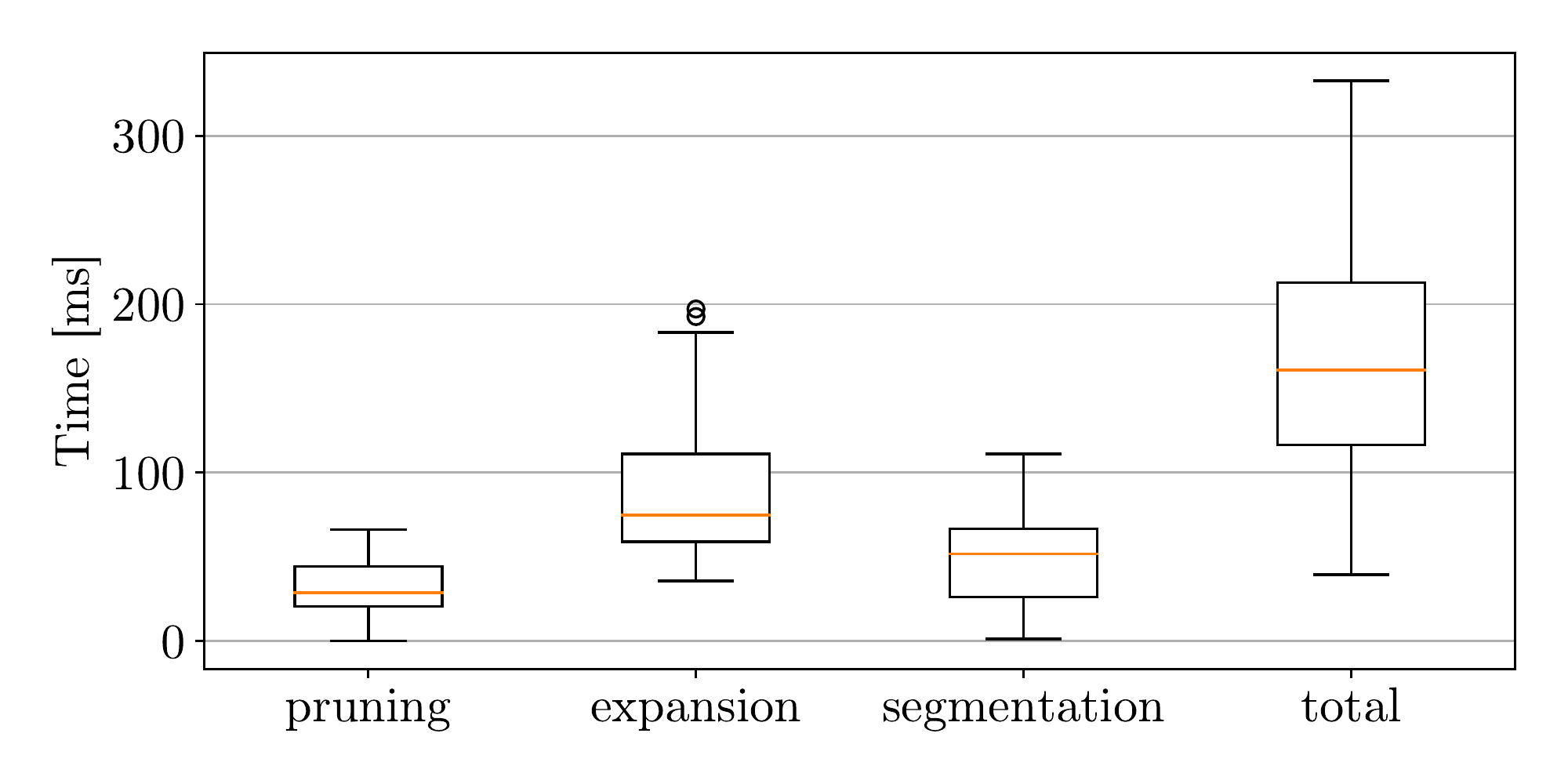}
  \caption{
    Boxplot illustration of runtimes of individual parts of one \map{} growing iteration during an exploration mission in a large cave environment. 
    The outlier computation times usually occur when the UAV peeks around a corner and discovers a vast free-space area.}
  \label{fig:spheremap_runtimes}
\end{figure}


\subsection{Map compression analysis}
\label{section:map_compression}
To show the usefulness and light weight of the \segmap{} generated from the \map{}, we conducted an experiment in a simulated cavernous environment spanning hundreds of meters and compared the \segmap{} with the original binary occupancy octree (\SI{0.2}{\meter} resolution), from which the \map{} is built.
We also compare the size of the \segmap{} and an occupancy octree that is generated by lowering the original octree to a resolution of \SI{1}{\meter}.
The development of the size of the three representations throughout the mission is shown in~\autoref{fig:map_compression_timelapse}. 
The \segmap{} and the original map at the end of the experiment are visualized in~\autoref{fig:map_compression_octomap}.

According to~\autoref{fig:map_compression_timelapse}, the lowered-resolution occupancy octree could seem like a good choice for volumetric information sharing.
However, the resolution lowering can cause narrow passages to become obstructed in the resulting maps, which could give other robots false information about the environment's connectivity.
The \segmap{} on the other hand keeps this connectivity information while also being smaller than the lowered-resolution map.

One disadvantage of the shown \segmap{} is that since we use 4DOF bounding boxes for the region shape approximation, there is a large percentage of space in the bounding boxes, that is not free in the original occupancy octree.
The amount of misclassified space could be reduced by choosing different bounding shapes, \vk{such as polygons or 6DOF boxes, which would better approximate the segments' shapes} at the expense of higher computational complexity and higher communication bandwidth.
Nevertheless, for the purposes of decentralized cooperative planning, we found the 4DOF boxes to be sufficient in the DARPA SubT Challenge~\vk{\cite{petrlik2022uavs}}.

\begin{figure}[H]
  \definecolor{blue}{HTML}{2078B4}
  \definecolor{orange}{HTML}{FF7F0E}
  \definecolor{green}{HTML}{2DA02D}
  \newcommand{\blueline}{\raisebox{2pt}{\tikz{\draw[blue,solid,line width = 1.0pt](0,0) -- (5mm,0);}}}
  \newcommand{\orangeline}{\raisebox{2pt}{\tikz{\draw[orange,solid,line width = 1.0pt](0,0) -- (5mm,0);}}}
  \newcommand{\greenline}{\raisebox{2pt}{\tikz{\draw[green,solid,line width = 1.0pt](0,0) -- (5mm,0);}}}
  \centering
  \adjincludegraphics[width=0.8\linewidth, trim={{0.00\width} {0.07\height} {0.00\width} {0.06\height}}, clip]{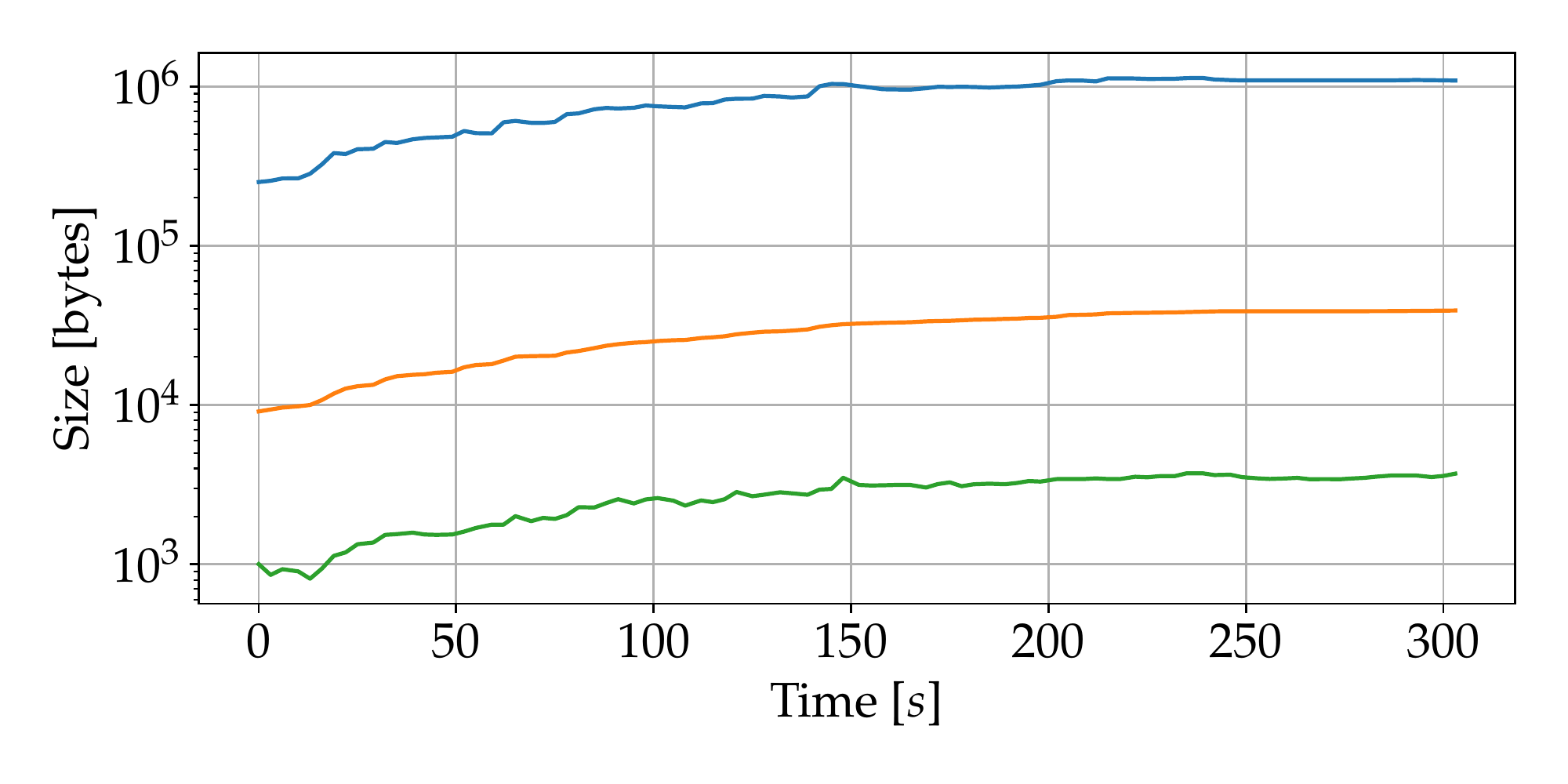}

  \caption{Development of the size of the \segmap{} (\protect\greenline), the binary occupancy octree (\protect\blueline) used by the \map{} (implemented by the OctoMap library \cite{octomap}), and an octree generated by lowering the original octree to a \SI{1}{\meter} resolution (\protect\orangeline) in the experiment described in~\autoref{section:map_compression}.}
  \label{fig:map_compression_timelapse}
\end{figure}

\begin{figure}[H]
  \centering
  \begin{tikzpicture}
    \centering
    \node[anchor=south west,inner sep=0] (a) at (0,0) {\adjincludegraphics[width=1.0\linewidth, trim={{0.0\width} {0.0\height} {0.0\width} {0.0\height}}, clip=true]{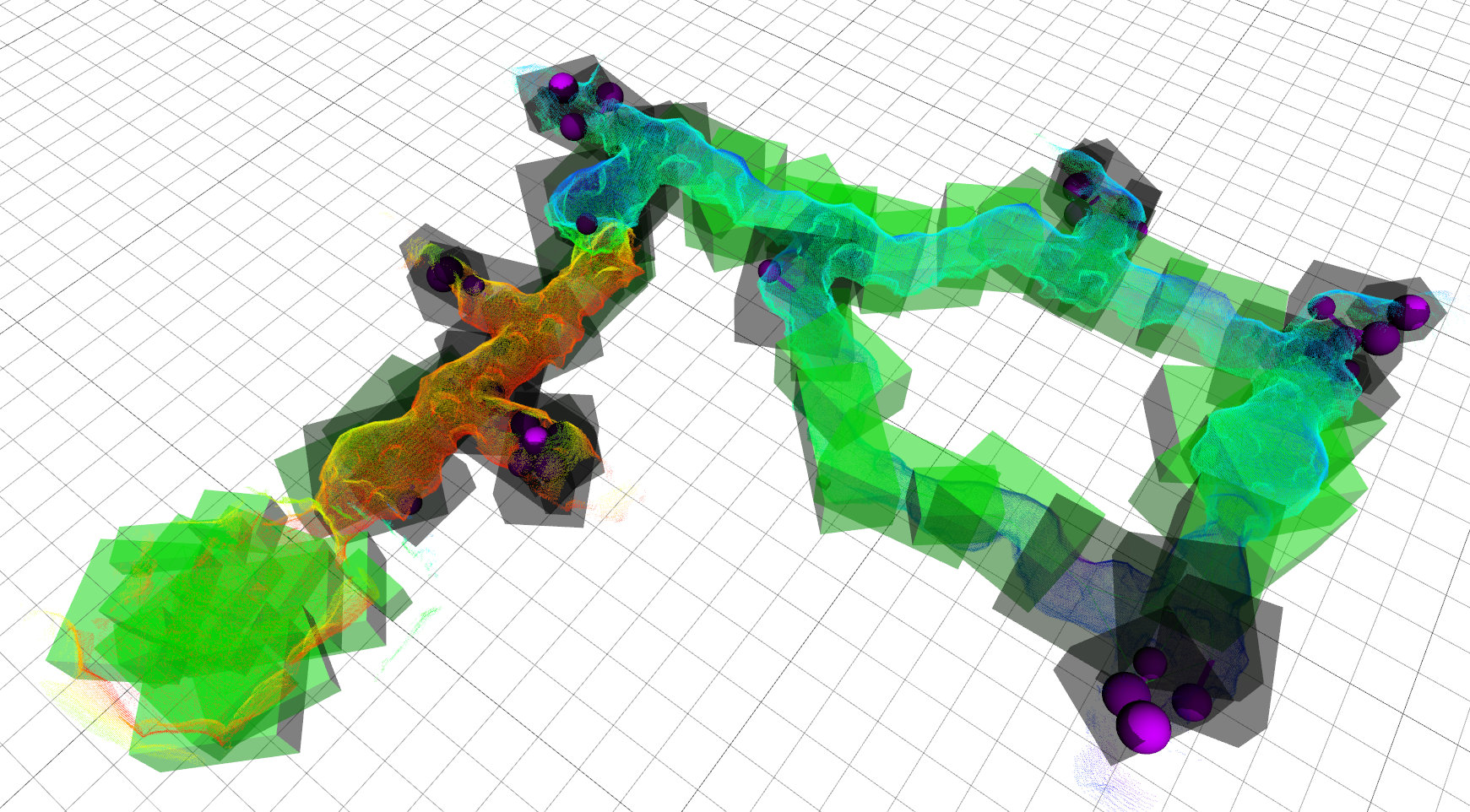}};
    \begin{scope}[x={(a.south east)},y={(a.north west)}]
      \draw [latex-latex](0.03,0.58) -- (0.243,0.855);
      \node[align=center] at (0.11, 0.77) {\scriptsize \color{black}\SI{100}{\meter}};
    \end{scope}
  \end{tikzpicture}

  \caption{Visualization of the original occupancy octree (pointcloud colored by a heatmap encoding the point height) and the \segmap{} generated from the \map{} built from it in the experiment described in~\autoref{section:map_compression}. The purple spheres signify potential exploration goals appended to the \segmap{} and the color of the blocks signifies how close they are to unexplored space.}
  \label{fig:map_compression_octomap}
\end{figure}


\subsection{Real world deployment}
The \map{} described in this \paper{} was successfully used in the final round of the DARPA SubT Challenge (\autoref{fig:intro_img}) for the purpose of navigation, cooperative exploration, goal pathfinding, and map sharing among robots and with an operator.
The control and state estimation part of the deployed system is described further in \cite{baca2021mrs}.
The \map{} in a more crude form (without pruning of nodes) was used also in the virtual circuit of the DARPA SubT Challenge Finals, where our team CTU-CRAS-NORLAB won the \nth{2} place~\vk{\cite{petrlik2022uavs}}.

Lastly, the \map\ was also tested on a large outdoor environment where a UAV autonomously explored the space around a house surrounded by heavy shrubbery, crossed a bridge into a vast open field, and safely returned home. 
With a minimal amount of fine-tuning (only lowering the number of sampled points to compensate the large free space volume), the total \map\ growth runtime was on average around \SI{200}{\milli\second} on a real-world UAV platform equipped with an Intel Core i7 processor running SLAM, state estimation, visual detection and control in addition to \map{}, which demonstrates that the proposed method is usable on common hardware even in large outdoor areas.
Video outputs from the outdoor exploration experiment and from both the virtual and physical DARPA SubT Final events are available in the supplementary materials.


\section{Conclusions}
A method for flexible in-flight building of a segmented graph of spheres with cached intra-segment paths designed for rapid safety-aware path planning and generating a lightweight topological-volumetric graph representation of the explored environment on-demand was proposed in this \paper{}.
The proposed method does not require any fine-tuning of the map resolution based on apriori knowledge of the environment.
Thus, for achieving safe and flexible multi-robot exploration of unknown unstructured environment, the proposed approach requires only a map of occupied, free and unknown space in the vicinity of the UAV that can be efficiently provided by the onboard sensors at every iteration of the map update.
Planning in the proposed \map{} is designed to find paths that optimize both safety and length.
The topological-volumetric \segmap{} generated from the proposed structure takes up significantly less bandwidth than the occupancy map even with a lowered resolution, while still keeping information about the connectivity of the environment and its approximate shape, which is crucial for real-time multi-robot exploration of subterranean environment, where reliable communication among robots is a bottleneck.

Among numerous simulations and statistical analyses using world models from scanned caves, mines and industrial complexes, the proposed method was also verified onboard real UAV platforms during multiple exploratory missions and was able to find paths 3 orders of magnitude faster than state-of-the-art planning algorithms such as grid-based A* or RRT*, navigate robots in the environment safely, and enable cooperation among the UAVs sharing the lightweight maps.
The performance of the entire UAV system using the proposed method was confronted with competition consisting of leading robotic teams selected for finals of DARPA SubT Challenge.
The challenge provided trustworthy conditions for comparison of different solutions in real-world conditions of demanding cave, mine and subway domains.
The proposed method successfully navigated a team of UAVs through dust-filled narrow tunnels with dynamically blocked paths in the systems track and was a crucial part of the system deployed in the virtual track, gaining the \nth{2} place.

\balance
\bibliographystyle{IEEEtran}
\bibliography{main.bib}

\end{document}